%% file: arx-ijcai.tex
\documentclass{article}
\usepackage{ijcai26}

 \usepackage{pifont}
 \usepackage{multirow}
 \usepackage{dblfloatfix}

\usepackage{algorithm}
\usepackage{algpseudocode}
\usepackage{array}

\usepackage{times}
\usepackage{soul}
\usepackage{url}
\usepackage[hidelinks]{hyperref}
\usepackage[utf8]{inputenc}
\usepackage[small]{caption}
\usepackage{graphicx}
\usepackage{amsmath}
\usepackage{amssymb}
\usepackage{amsthm}
\usepackage{booktabs}
\usepackage{xcolor}
\usepackage[switch]{lineno}
\usepackage{subcaption}
\usepackage{adjustbox} 

\usepackage[T1]{fontenc}
\usepackage{multirow}
\usepackage{makecell}
\usepackage{tabularx}
\usepackage{threeparttable}
\usepackage{colortbl}
\usepackage{comment}

\usepackage{pifont}
\newcommand{\cmark}{\ding{51}}
\newcommand{\xmark}{\ding{55}}
\newcommand{\rot}[1]{\rotatebox{90}{\hspace{1pt}#1\hspace{1pt}}}

\title{PAR: Plausibility-aware Amortized Recourse Generation}

\author{
    Anagha Sabu, Vidhya S, Narayanan C Krishnan
    \affiliations
    Mehta Family School of Data Science and Artificial Intelligence, IIT Palakkad
    \emails
    \{142404001,142414002\}@smail.iitpkd.ac.in, ckn@iitpkd.ac.in
}

\begin{document}

\onecolumn
\maketitle

\begin{abstract}
Algorithmic recourse aims to recommend actionable changes to a factual's attributes that flip an unfavorable model decision while remaining realistic and feasible. We formulate recourse as a Constrained Maximum A-Posteriori (MAP) inference problem under the accepted-class data distribution seeking counterfactuals with high likelihood while respecting other recourse constraints. We present PAR, an amortized approximate inference procedure that generates highly likely recourses efficiently. Recourse likelihood is estimated directly using tractable probabilistic models that admit exact likelihood evaluation and efficient gradient propagation that is useful during training. The recourse generator is trained with the objective of maximizing the likelihood under the accepted-class distribution while minimizing the likelihood under the denied-class distribution and other losses that encode recourse constraints. Furthermore, PAR includes a neighborhood-based conditioning mechanism to promote recourse generation that is customized to a factual. We validate PAR on widely used algorithmic recourse datasets and demonstrate its efficiency in generating recourses that are valid, similar to the factual, sparse, and highly plausible, yielding superior performance over existing state-of-the-art approaches.

\end{abstract}

\section{Introduction}
\label{sec:intro}
Machine learning models increasingly inform decisions in socially and economically significant domains \cite{Wachter2018CounterfactualGDPR,Karimi2022AlgorithmicRecourse}, including credit scoring, hiring, healthcare triage, and public policy. As such systems become more pervasive, individuals subject to unfavorable predictions require mechanisms that go beyond post-hoc explanations and instead provide actionable guidance on how an outcome might be changed. \textit{Algorithmic recourse} addresses this need by recommending minimal modifications to a user’s attributes that would flip an undesirable prediction to a desirable one, akin to counterfactual explanations \cite{Ustun2019ActionableRecourse,Karimi2022AlgorithmicRecourse}. Crucially, for recourse to be meaningful in practice, these recommendations must be feasible, respect immutable constraints, and consistent with the underlying data-generating process; failing which risks producing implausible suggestions undermining user trust.

Despite substantial interest, existing methods for generating algorithmic recourse exhibit several limitations. Many approaches rely on local, gradient-based perturbations \cite{Mothilal2020DiCE,VanLooveren2019ProtoCF,Joshi2019REVISE,Poyiadzi_2020,Kugelgen2022FairCausalRecourse}  or proximity-driven search in feature space, with limited consideration of the global structure of the underlying data distribution. As a consequence, such methods may recommend changes that lie off the data manifold or violate implicit dependencies between features. While some recent works attempt to promote plausibility through regularization terms \cite{Karimi2022AlgorithmicRecourse,Guidotti2022Counterfactual} or post-hoc filtering, these mechanisms are heuristic and do not account for the explicit probabilistic data model. Consequently, the resulting recourse suggestions, though effective at altering predictions, may be unrealistic for individuals in practice.

To address these challenges, we adopt a generative modeling perspective and cast algorithmic recourse as a constrained maximum-a-posteriori (MAP) inference problem. Specifically, we seek counterfactual instances that are both likely under the data distribution and sufficient to flip an unfavorable prediction, while respecting immutable feature constraints and minimizing deviation from the factual instance. To model the data distribution tractably, we leverage \textit{probabilistic circuits} (PC), a class of deep generative models that admit efficient and exact likelihood computation as well as conditional inference \cite{Choi2020ProbabilisticCircuits,Darwiche2003Differential}. Within this formulation, plausibility is enforced directly through the generative model, instead of an external regularizer. Solving the resulting constrained MAP problem exactly is computationally challenging; therefore, we introduce an \textit{amortized approximate inference strategy} that learns to generate high-quality recourse efficiently.
\begin{figure}[t]
    \centering
    \includegraphics[width=1\linewidth]{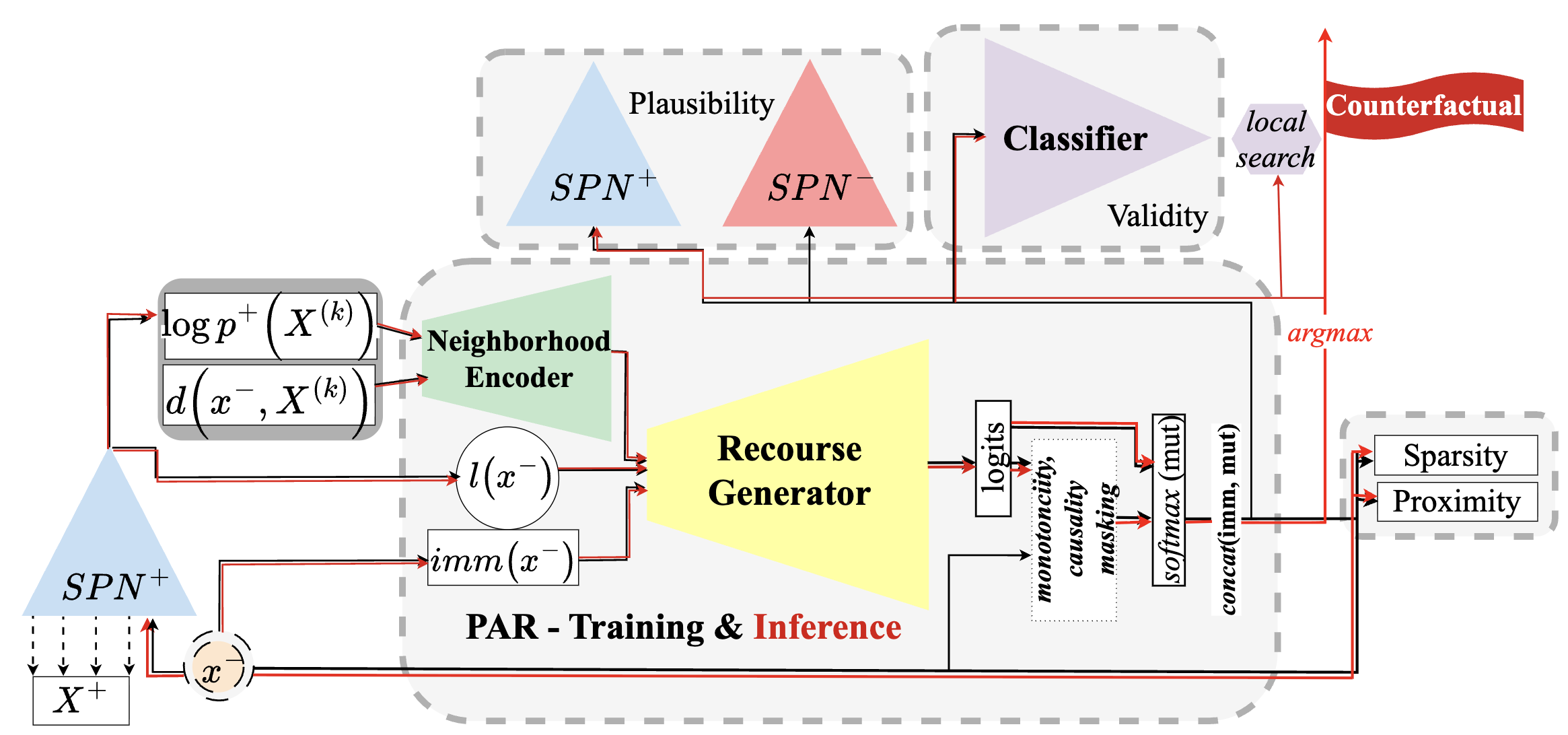}
    \caption{Overview of PAR shows the PAR architecture, where class-conditional probabilistic circuits guide a neural recourse generator to produce valid and plausible counterfactuals under structural constraints.}
    \label{fig:PAR_redesigned}
\end{figure}


Overall, we make the following contributions:
\begin{itemize}
    \item We formulate recourse generation as a constrained MAP inference problem over PC and develop an efficient approximate inference procedure for solving it.
    \item Our framework integrates plausibility as the optimality criterion and other recourse constraints as the feasibility criterion, addressing a key deficiency in existing recourse methods.
    \item We validate the proposed approach on widely used benchmark datasets, demonstrating superior performance relative to the state-of-the-art baselines.
\end{itemize}

\section{Related Work}
\label{sec:relatedwork}
Building on counterfactual explanations, \cite{Karimi2022AlgorithmicRecourse} present a comprehensive survey of algorithmic recourse, which is concerned with providing actionable explanations. \cite{Guidotti2022Counterfactual} comprehensively reviews and benchmarks existing methods for generating counterfactual explanations, categorizing algorithms such as DiCE, FACE, and C-CHVAE against key desiderata including plausibility, validity, proximity, actionability, and diversity. Similar approaches in the search of counterfactual explanation such as \cite{Ustun2019ActionableRecourse,Wachter2018CounterfactualGDPR}, focus on the objectives related to plausibility. 

Several prior works incorporate plausibility into recourse generation, but they differ substantially in how it is defined and enforced. VAE-based approaches such as C-CHVAE \cite{Pawelczyk_2020} learn latent representations of tabular data to produce sparse and valid counterfactuals, thereby encouraging plausibility. However, plausibility is only assessed indirectly: likelihoods are inexact, and generated counterfactuals are post hoc checked for outlierness rather than optimized under an explicit density model. BETRACE \cite{St_pka_2025} is another model-agnostic framework that provides probabilistic robustness guarantees for counterfactuals, but does not directly model data density. FACE \cite{Poyiadzi_2020} relies on kernel density estimation to encourage plausibility, which scales poorly to high-dimensional settings and handles categorical variables inadequately. DiCE \cite{Mothilal2020DiCE} formulates counterfactual generation as a constrained optimization problem emphasizing diversity, but does not explicitly enforce plausibility through a data distribution. PROPLACE \cite{jiang2024provablyrobustplausiblecounterfactual} incorporates plausibility via feasibility constraints—such as convex-hull membership—yet this geometric criterion can admit low-density outliers and does not account for relative data likelihoods.

Most closely related to our work is LiCE \cite{nemecek2025generatinglikelycounterfactualsusing}, which formulates plausible counterfactual generation as a mixed-integer optimization problem. Feature domains are modeled using mixed polytopes, and plausibility is enforced via the log-likelihood of a Sum-Product Network. While this unified MIO–SPN formulation yields valid and distributionally plausible counterfactuals, it incurs substantial computational overhead: discretization, categorical encodings, ReLU linearizations, and SPN sum-node approximations introduce many binary variables and big-M constraints, limiting scalability and introducing approximation error.

\section{Preliminaries}
\label{sec:prelims}
\subsection{Notation}
\label{sec:notation}
Let $\mathcal{X} \subset \mathbb{R}^D$ denote the input space, where $\mathbf{x} = (x_1,\ldots,x_D)$ and $x_j$ is the $j$-th feature. We consider a differentiable classifier $ f : \mathcal{X} \rightarrow [C],$ trained on a dataset $\mathcal{D} = \{(\mathbf{x}_i, y_i)\}_{i=1}^N$ with $y_i \in [C] := \{1,\ldots,C\}$. Although our method extends to the multi-class setting, we focus on binary classification with labels $\{0,1\}$. We denote the class-conditional distributions of the positive and negative classes by $p^+$ and $p^-$, respectively.

\subsection{Recourse}
\label{sec:recourse}
Following prior work \cite{Wachter2018CounterfactualGDPR,Ustun2019ActionableRecourse,Mothilal2020DiCE,Karimi2022AlgorithmicRecourse,Guidotti2022Counterfactual,Pawelczyk_2020,Karimi2021CausalRecourse}, a \emph{factual} instance $\mathbf{x}^- \in \mathcal{X}$ yields an undesired prediction, i.e., $f(\mathbf{x}^-)= 0$. A \emph{counterfactual} (or recourse) is an instance $\mathbf{x}^+ \in \mathcal{X}$ satisfying $f(\mathbf{x}^+) = 1$. In addition to achieving label validity, a desirable counterfactual $\mathbf{x}^+$ is expected to satisfy:
\begin{itemize}
    \item \textit{Proximity:} $\mathbf{x}^+$ is close to $\mathbf{x}^-$ under a suitable metric.
    \item \textit{Sparsity:} Only a small number of features are modified.
    \item \textit{Actionability:} \textit{Immutable} features remain fixed and domain constraints (e.g., monotonicity) are respected.
    \item \textit{Plausibility:} $\mathbf{x}^+$ has high likelihood under $p^+$.
    \item \textit{Causality:} Feature changes are consistent with known causal relations.
\end{itemize}

\subsection{Probabilistic Circuits}
\label{sec:pcs}
A \textbf{Probabilistic Circuit} (PC) $p$ is a parameterized directed acyclic graph (DAG) \cite{Darwiche2020ProbabilisticCircuits} with a unique root node $n_r$ that represents a joint distribution over random variables $\mathbf{X}=\{X_1,\ldots,X_D\}$, with realizations $\mathbf{x}\in\mathcal{X}\subset\mathbb{R}^D$. The DAG consists of three node types: input nodes representing univariate distributions $f_n$, sum nodes computing convex combinations and product nodes computing products of their children's outputs respectively.

Each node $n$ computes a function $p_n(\mathbf{x})$ recursively as
\[
p_n(\mathbf{x}) =
\begin{cases}
f_n(\textbf{x}), & \text{if $n$ is an input node}, \\
\prod_{c \in in(n)} p_c(\mathbf{x}), & \text{if $n$ is a product node}, \\
\sum_{c \in in(n)} \theta_{nc}\, p_c(\mathbf{x}), & \text{if $n$ is a sum node},
\end{cases}
\]
where $in(n)$ denotes the children of $n$ and $\theta_{nc}\in(0,1]$ satisfy $\sum_{c\in in(n)}\theta_{nc}=1$.

Each node $n$ is associated with a \emph{scope} $\phi(n)\subseteq\mathbf{X}$; for internal nodes, $\phi(n)=\bigcup_{c\in in(n)}\phi(c)$, where $in(c)$ represents the children of $n$, and the root has scope $\mathbf{X}$. Sum nodes thus represent mixtures over identical scopes, while product nodes represent factorizations over disjoint scopes.

PCs are evaluated bottom-up, with the joint distribution given by $p(\mathbf{x})=p_{n_r}(\mathbf{x})$. We assume alternating layers of sum and product nodes. For tractable inference \cite{Poon2011SPN,Peharz2017SPN}, PCs are typically required to be \emph{smooth} (children of each sum node have identical scopes) and \emph{decomposable} (children of each product node have disjoint scopes). Throughout this work we assume that the learned PCs are smooth and decomposable.


\section{PAR - Recourse as Constrained MAP Inference}
\label{sec:par}

Framing recourse as a constrained MAP inference problem over the positive-class data distribution directly enforces plausibility. Rather than relying on heuristic penalties or surrogate objectives, this approach seeks the most probable feasible instance under $p^+$, ensuring that counterfactuals lie in high-density regions of the desired outcome distribution. Recourse desiderata such as proximity, sparsity, actionability, and monotonicity can be naturally expressed as constraints, yielding a clean separation between data-driven modeling and domain-specific requirements. Moreover,  this formulation avoids exploiting classifier boundary artifacts and produces recourses that are statistically robust.

Formally, let $\mathbf{x}^- \in \mathcal{X}$ be a factual instance such that
$f(\mathbf{x}^-)=0$, where $f:\mathcal{X}\to\{0,1\}$ denotes a fixed, deployed
binary classifier and $f(\mathbf{x})=1$ corresponds to the desired outcome.
The goal of PAR is to compute a recourse
$\mathbf{x}^+\in\mathcal{X}$ that flips the classifier decision while remaining
plausible and feasible. We formulate recourse as constrained MAP inference under the
accepted-class distribution $p^+$:
\begin{equation}
\mathbf{x}^+ \;=\; \arg\max_{\mathbf{x}\in\mathcal{X}} \;\log p^+(\mathbf{x})
\quad \text{s.t.} \quad f(\mathbf{x}) = 1,
\label{eq:PAR-map}
\end{equation}
\begin{equation}
\mathbf{x} \in \mathcal{F}(\mathbf{x}^-),
\label{eq:PAR-feasible}
\end{equation}
where $\mathcal{F}(\mathbf{x}^-)$ denotes the feasible set encoding recourse
requirements, including proximity to $\mathbf{x}^-$, sparsity of feature
changes, causal constraints and actionability constraints such as immutability and monotonicity.
 We assume access only to the differentiable classifier
$f$ at test time, and not to its training data. PAR realizes this objective through a three-phase pipeline: \emph{(i) Pretraining}, \emph{(ii) Generator Training}, and \emph{(iii) Local Search Refinement},
as illustrated in Figure~\ref{fig:PAR_redesigned}. PAR involves five components: a fixed classifier $f$, two fixed class-conditional generative models ($p^+$, $p^-)$, a neighborhood encoder $\mathcal{E}$ and a recourse generator $\mathcal{G}$. The classifier ($f$) and generative models $(p^+$, $p^-)$ are assumed to have been learned using the training data apriori and training PAR involves learning $\mathcal{E}$ and $\mathcal{G}$ only.

\subsection{Phase I: Pretraining}
\label{sec:phaseone}
\paragraph{Categorical Representation via Discretization.}
PAR discretizes continuous features into quantile-based bins, yielding a fully categorical representation. Details of the same is provided in the Appendix \ref{app:discretization}. This produces actionable interval-valued recourses while enabling tractable plausibility optimization using PCs with categorical leaves, which support exact multilinear evaluation and gradients with respect to soft categorical inputs.

\subsubsection{Pretraining Components}
\label{sec:pretrainingcomponents}

We use a differentiable classifier $f$ to enable end-to-end training of the neighborhood encoder $\mathcal{E}$ and the generator $\mathcal{G}$. Although the class-conditional densities $(p^+, p^-)$ can in principle be modeled using various generative models, we adopt PC—specifically Sum-Product Networks—because they naturally handle categorical features and support exact, tractable log-likelihood computation. Alternative models such as GMMs, VAEs, and GANs either struggle with categorical variables or do not provide exact likelihoods. While the MAP objective depends only on $p^+$, the denied-class density $p^-$ is used as a contrastive signal to discourage counterfactuals that are simultaneously typical under both classes. Thus at the end of phase I, we have access to the learned classifier $f$ and the PCs $p^+$ and $p^-$.

\subsection{Phase II: Amortized Recourse Generation}
\label{sec:phasetwo}
Solving Equation  ~\ref{eq:PAR-map} independently for each factual instance is computationally expensive. We therefore learn an amortized mapping from factual instances to approximate MAP solutions. We assume that the training data used to learn $f$, $p^+$ and $p^-$ are inaccessible beyond Phase I. \textit{The recourse generator is trained using instances sampled from $p^+$ and $p^-$.}

\subsubsection{Neighborhood Encoder}
\label{sec:ne}
 Multiple denied factual instances can share identical immutable feature values but require different changes because the feasible accepted region depends on the local geometry around $\mathbf{x}^-$. Hence, conditioning the generator only on immutable features is insufficient. We therefore introduce a neighborhood encoder that summarizes the local context around $\mathbf{x}^-$ using nearby accepted examples.

 Let $A$ denote a pool of accepted instances used for neighborhood construction obtained by sampling from $p^+$ and filtered by the fixed classifier: $A = \{\mathbf{x} \sim p^+: f(\mathbf{x}) = +\}$. We precompute $\log p^+(\mathbf{x})$ for all $\mathbf{x} \in A$ using $p^+$.

Given a denied factual $\mathbf{x}^-$, we estimate its K- nearest neighbors in $A$ using Hamming distance and define a descriptor that captures the distance and the likelihood of a nearby accepted instance. Each neighbor $x^{(k)}$ is represented as $ u^{(k)} = \left( d(\mathbf{x}^-, \mathbf{x}^{(k)}), \log p^+(\mathbf{x}^{(k)})\right)$. The Hamming distance term $d$ captures how far the factual is from a nearby accepted point and the log-likelihood term captures how typical that neighbor is under the accepted distribution. 
The neighborhood encoder aggregates these descriptors via
$ h = \mathcal{E}(\{u^{(k)}\}_{k=1}^K)
= \rho\!\left(\frac{1}{K}\sum_{k=1}^K \psi(u^{(k)})\right)
$, where \( \psi \) is a shared MLP and \( \rho \) is a linear projection.


   



 \subsubsection{Recourse Generator}
 \label{sec:generator}
Let $\mathcal{I} \subseteq \{1, ..., D\}$ be immutable features and $\mathcal{M} = \{1, ..., D\}\setminus \mathcal{I}$ be the mutable feature set. The recourse generator $\mathcal{G}$ is modeled as a MLP that is conditioned on: (i) factual immutable features, (ii) the factual accepted-likelihood score $\log p^+(\mathbf{x}^-)$, and (iii) the neighborhood score $h$. Generator takes $
    \mathbf{z} = \left[(\mathbf{x}^-_\mathcal{I}), \log p^+(\mathbf{x}^-), h\right]$ as input where $\mathbf{x}^-_\mathcal{I}$ is the concatenated immutable features, and outputs $\mathbf{l}=\mathcal{G}(\mathbf{z})
$, the logits for all mutable variables. This neighborhood conditioning allows two factuals with identical immutables to yield different recourse distributions when their local accepted neighborhoods differ.

 \par Rather than directly outputting a hard recourse, the generator produces feature-wise categorical logits for all mutable attributes. After applying a softmax independently per feature, this yields a set of categorical distributions $q_j(\cdot|x^-)$, where $j \in \mathcal{M}$.
 This makes $\mathcal{G}$ fully differentiable, enabling end-to-end training under multiple objectives. It allows feasibility constraints such as monotonicity and causality at the logits level to enforced them directly at the probability mass level. During inference, a hard recourse is obtained by taking the per-feature argmax.

 \par This design also enables flexibility under changing assumptions. When the immutable feature set is fixed, a single trained generator can be reused indefinitely. A new generator can be trained for a different set of immutable features without access to the original training data, using only samples from the pretrained PCs. 

\subsubsection{Modeling the Recourse Feasibility Constraints}
\label{sec:feasconstraintsmodeling}
\par \textbf{\textit{Monotonicity}} and \textbf{\textit{causal}} recourse constraints are enforced deterministically via logit masking. This guarantees feasibility while preserving differentiability. Monotonic features are restricted to non-decreasing changes, and causal dependencies are enforced either through joint masking (when both cause and effect features are mutable) or single-feature masking (when the cause is immutable). Treating these as hard constraints ensures that every generated recourse strictly respects the prescribed monotonicity and causal relations. Please refer to the Appendix \ref{app:constraints} for more details.

\par \textbf{\textit{Validity:}}
Concretely, we evaluate the fixed classifier on the soft counterfactual representation $\mathbf{q}$ (a concatenation of immutable blocks clamped to the factual and mutable categorical distributions). We then apply binary cross entropy ($BCE$) against the positive label: $L_{val} = BCE(f(\mathbf{q}), 1)$. 

\par \textbf{\textit{Proximity:}}
Proximity is treated as a feasibility requirement that restricts the admissible search region of the MAP problem. Rather than enforcing this constraint combinatorially, we impose it in expectation under the generator’s outputs.

For each mutable feature \( j \in \mathcal{M} \), the generator produces a categorical distribution \( q_j(\cdot) \). The probability of modifying feature \( j \) relative to the factual instance \( \mathbf{x}^- \) is $ \pi_j \;=\; 1 - q_j(x^-_j)$. The expected number of modified mutable features, i.e., the expected Hamming distance, is then $d_{\mathrm{ham}}(\mathbf{x}^-, \mathbf{q}) \;=\; \sum_{j \in \mathcal{M}} \pi_j.$ Ideally, proximity would be enforced as a hard feasibility constraint of the form $d_{\mathrm{ham}}(\mathbf{x}^-, \mathbf{q}) \le B,$ thereby defining a restricted feasible set for the constrained MAP problem. In practice, as recourses are generated via an amortized model, we enforce this constraint in expectation using a soft penalty. Specifically, we employ a squared hinge loss
\begin{equation}
L_{\mathrm{prox}}
\;=\;
\max\!\left(0,\; \mathbb{E}\!\left[d_{\mathrm{ham}}(\mathbf{x}^-, \mathbf{q})\right] - B \right)^2,
\end{equation}
penalizing violations above the expected budget \( B \).


\par \textbf{\textit{Sparsity:}}
While the expected Hamming budget constrains the \emph{number} of feature changes in expectation, it does not prevent the generator from remaining diffuse by assigning nontrivial probability mass to many alternative edits. Such distributional indecision degrades interpretability after decoding and can interfere with the reliable enforcement of hard feasibility constraints. To address this, we introduce a sparsity regularizer that complements the proximity constraint by encouraging confidence in unchanged features. Specifically, we penalize deviation from the factual category using a stay-close cross-entropy term: $L_{\mathrm{sparse}}
\;=\;
\frac{1}{|\mathcal{M}|}
\sum_{j \in \mathcal{M}}
- \log q_j(x^-_j).$
Minimizing \( L_{\mathrm{sparse}} \) increases the probability mass assigned to the factual value for most mutable features, concentrating the distribution on ``no change'' unless validity or plausibility necessitates an edit. In contrast to proximity—which controls expected sparsity—this term sharpens the generator’s output distribution, resulting in a small subset of features that are modified with high confidence.



\par \textbf{\textit{Entropy:}}
Recourse generation ultimately requires a single discrete counterfactual, obtained by hard decoding of the generator’s output. High-entropy outputs can lead to unstable decoding behavior, where small perturbations result in qualitatively different counterfactuals and inconsistent satisfaction of feasibility constraints. We therefore encourage decisiveness at the distributional level by penalizing entropy. For each mutable feature \( j \in \mathcal{M} \), we add an entropy regularizer $L_{\mathrm{ent}}
\;=\;
\frac{1}{|\mathcal{M}|}
\sum_{j \in \mathcal{M}}
\left(
-\sum_{a=0}^{C_j-1}
q_{j,a}\log q_{j,a}
\right).$

Minimizing \( L_{\mathrm{ent}} \) promotes low-entropy, near one-hot distributions, ensuring that the amortized MAP approximation yields stable and deterministic counterfactuals after decoding. This term complements the sparsity regularizer: while \( L_{\mathrm{sparse}} \) biases the generator toward retaining the factual category, the entropy penalty encourages confidence in whichever category is selected. Together, these terms sharpen the generator’s output distribution, improving interpretability and robustness without introducing additional feasibility constraints.


\par \textbf{\textit{Plausibility:}}
Plausibility is treated as the optimality criterion of the constrained MAP formulation. Among recourses that satisfy validity and proximity-based feasibility constraints, we seek solutions that are typical of the accepted population and atypical of the denied population. This is enforced directly through tractable log-likelihoods provided by PC, evaluated on the generator’s soft counterfactual encoding. We introduce an attraction term toward the accepted-class density $L_{+}
\;=\;
-\,\mathbb{E}\!\left[\log p^{+}(\mathbf{q})\right],$ and a repulsion term from the denied-class density $L_{-}
\;=\;
\mathbb{E}\!\left[\log p^{-}(\mathbf{q})\right].$
Minimizing \( L_{+} \) increases likelihood under the accepted distribution, while minimizing \( L_{-} \) discourages regions that are typical of the denied population. Together, these terms implement a relative plausibility objective consistent with MAP inference under competing class-conditional models.

\paragraph{Total Objective.}
The full training loss combines feasibility-enforcing terms with MAP optimality terms:
\begin{align}
L
&=
\lambda_{\mathrm{val}}\, L_{\mathrm{val}}
+
\lambda_{\mathrm{ppt}}
\Big(
\alpha\, L_{\mathrm{prox}}
+
(1-\alpha)\big(\lambda_{+}L_{+}+\lambda_{-}L_{-}\big)
\Big)
\nonumber\\
&\quad+
\lambda_{\mathrm{sparse}}\,L_{\mathrm{sparse}}
+
\lambda_{\mathrm{ent}}\,L_{\mathrm{ent}} .
\end{align}
The scalar $\alpha \in [0,1]$ controls the relative weighting between proximity and plausibility within the objective.
The generator thus learns an amortized approximation to the constrained MAP solution. A discrete counterfactual is then obtained by decoding,
\( x^{+}_j = \arg\max_a q_{j,a} \),
yielding a stable initialization that lies near a feasible MAP solution.

\textbf{Training the Recourse Generator} requires the gradient from the PCs $p^+$ and $p^-$; which can be computed in time linear in the size of the PC. Let the categorical feature $X_j$ take values in $\{0,\ldots,C_j-1\}$. We evaluate the circuits on soft categorical inputs $q=(q_1,\ldots,q_D)$, with each $q_j \in \Delta^{C_j-1}$. Leaf nodes compute multilinear expectations $l_j(q_j)=\sum_{c=0}^{C_j-1} q_{j,c}\theta_{j,c}$, while internal sum and product nodes follow standard PC semantics. Consequently, the circuit output $p^+(q)$ (or $p^-$) is multilinear in each block $q_j$. Let $e_c$ denote the one-hot vector for category $c$, and let $q_{j\leftarrow e_c}$ denote the soft input obtained by replacing $q_j$ with $e_c$. By multilinearity, $p^+(q)=\sum_{c=0}^{C_j-1} q_{j,c}\, p^+(q_{j\leftarrow e_c})$, which implies $\partial p^+(q)/\partial q_{j,c} = p^+(q_{j\leftarrow e_c})$. Therefore, for smooth and decomposable PCs with categorical leaves, the gradient $\nabla_q \log p^+(q)$ can be computed exactly via bottom-up circuit evaluations: for each feature $j$, all partial derivatives $\{\partial \log v_{\mathcal{C}}(q)/\partial q_{j,c}\}_{c=0}^{C_j-1}$ are obtained using $C_j$ circuit evaluations, each linear in the size of $p^+$. Please refer to the Appendix \ref{app:pc_grad} for the detailed proof.

\subsection{Phase III: Local Search Refinement}
\label{sec:phasethree}
As the generator provides an amortized approximation rather than an exact solution to the constrained MAP problem, PAR applies a lightweight local refinement step at inference time. This refinement operates entirely within the feasible set \(\mathcal{F}(\mathbf{x}^-)\), strictly preserving immutability, monotonicity, and causal constraints. To remain consistent with the learned proximity objective, the search is restricted to candidates whose Hamming distance over mutable features does not exceed that of the generator’s decoded output.

If the decoded counterfactual already satisfies the classifier constraint, local refinement performs greedy sparsification, removing superfluous feature changes while maintaining feasibility. Otherwise, it performs constrained single-feature exploration within the fixed Hamming budget to repair validity. Among feasible candidates, priority is given to valid solutions, with ties broken by sparsity and positive-class likelihood \(\log p^{+}(\mathbf{x})\). Once validity is achieved, sparsification is applied again under the same constraints. Full algorithmic details are provided in Appendix~\ref{app:local_search}.

\begin{table*}[h!]
\centering
\caption{
Actionable counterfactual results on Adult, Credit, and GMSC.
Higher is better for \emph{Validity}, \emph{Actionability}, and \emph{Causality}.
Lower is better for \emph{Plausibility (NLL)}, \emph{Similarity}, \emph{Sparsity}, and \emph{Time}.
Percentages are reported as mean $\pm$ std when available (PAR).
Other metrics are mean $\pm$ std. Time is median seconds (baselines from LiCE timing table; PAR from our timing report).
Methods above the double rule operate in raw feature space; PAR methods operate in discretized space. Validity and plausibility are computed with respect to the model defined on each method’s operating space: validity is evaluated using the classifier defined on that space, and plausibility using the corresponding density model (SPN); this corresponds to raw feature space for baseline methods and discretized feature space for PAR.
}
\label{tab:recourse_main}

\resizebox{\textwidth}{!}{%
\small
\setlength{\tabcolsep}{5pt}
\renewcommand{\arraystretch}{1.15}
\begin{tabular}{lccccccc}
\toprule

\multicolumn{8}{c}{\textbf{Adult} \cite{BeckerKohavi1996Adult}} \\
\midrule
Method &
Validity (\%) &
Actionability (\%) &
Causality (\%) &
Plausibility (NLL)$\downarrow$ &
Similarity$\downarrow$ &
Sparsity$\downarrow$ &
Time (s)$\downarrow$ \\
\midrule
VAE            & \textbf{100.00} & 24.08 & 82.56 & 18.35 $\pm$ 0.40 & 19.16 $\pm$ 2.57 & 5.03 $\pm$ 0.20 & 0.920s \\
CVAE           & 16.72  & 11.26 & 96.35 & 18.00 $\pm$ 0.74 & \textit{4.07 $\pm$ 0.68}  & \textbf{2.44 $\pm$ 0.21} & 0.660s \\
DiCE           & 99.75  & 47.01 & 83.68 & 21.27 $\pm$ 0.45 & 9.10 $\pm$ 0.61  & 4.60 $\pm$ 0.18 & 18.40s \\
FACE(knn)      & \textbf{100.00} & 56.95 & 91.25 & 14.52 $\pm$ 0.37 & 4.78 $\pm$ 0.45  & 3.48 $\pm$ 0.10 & 7.120s \\
FACE($\epsilon$)& 79.12 & 38.94 & 88.63 & 14.90 $\pm$ 0.31 & 5.04 $\pm$ 0.61  & 3.56 $\pm$ 0.13 & 7.250s \\
PROPLACE       & \textbf{100.00} & 68.71 & 89.07 & 15.40 $\pm$ 0.93 & 6.40 $\pm$ 2.26  & 4.68 $\pm$ 0.26 & \textit{0.350s} \\
LiCE  & 91.96  & \textit{91.96}  & \textbf{100.00} & \textit{12.85 $\pm$ 0.07} & \textbf{1.68 $\pm$ 0.27} & \textit{2.90 $\pm$ 0.04} & 17.70s \\
\midrule\midrule
\textbf{PAR } & 93.59 & \textbf{100.00} & \textbf{100.00} & 11.76 $\pm$ 1.85 & 9.23 $\pm$ 1.02 & 4.75 $\pm$ 0.27 & \textbf{0.195s} \\
\textbf{PAR (+LS)}  & 98.45  & \textbf{100.00}  & \textbf{100.00}  & \textbf{11.27 $\pm$ 0.70} & 6.43 $\pm$ 1.37  & 3.26 $\pm$ 0.37 & 1.134s \\
\midrule

\multicolumn{8}{c}{\textbf{Credit} \cite{Hofmann1994GermanCredit}} \\
\midrule
Method &
Validity (\%) &
Actionability (\%) &
Causality (\%) &
Plausibility (NLL)$\downarrow$ &
Similarity$\downarrow$ &
Sparsity$\downarrow$ &
Time (s)$\downarrow$ \\
\midrule
VAE            & \textbf{100.00} & 0.00 & 78.12 & 48.30 $\pm$ 2.59 & 7.75 $\pm$ 0.38 & 10.53 $\pm$ 0.58 & 0.670s \\
CVAE           & 41.52  & 34.25 & 86.09 & 32.36 $\pm$ 1.35 & 7.35 $\pm$ 0.59 & 6.21 $\pm$ 0.33 & 0.560s \\
DiCE           & \textbf{100.00} & 0.62 & 72.84 & 35.95 $\pm$ 0.62 & 7.04 $\pm$ 0.72 & 7.13 $\pm$ 0.42 & 145.21s \\
FACE(knn)      & \textbf{100.00} & 39.82 & 84.74 & 44.26 $\pm$ 4.32 & 7.87 $\pm$ 0.50 & 6.74 $\pm$ 0.32 & 5.170s \\
FACE($\epsilon$)& 99.38 & 36.90 & 81.22 & 43.19 $\pm$ 1.34 & 7.63 $\pm$ 0.52 & 6.75 $\pm$ 0.25 & 5.080s \\
PROPLACE       & \textbf{100.00} & 33.85 & 87.91 & 38.15 $\pm$ 6.61 & 8.14 $\pm$ 0.79 & 8.64 $\pm$ 0.48 & \textit{ 0.290s} \\
LiCE  & 99.13 & \textit{99.13} & \textbf{100.00} & \textit{29.54 $\pm$ 0.68} & \textbf{0.47 $\pm$ 0.06} & \textbf{1.22 $\pm$ 0.13} & 2.930s \\
\midrule\midrule
\textbf{PAR} & 95.44  & \textbf{100.00 } & \textbf{100.00} & \textbf{18.18 $\pm$ 1.27} & 10.49 $\pm$ 1.81 & 6.92 $\pm$ 0.52 & \textbf{0.092s} \\
\textbf{PAR (+LS)}  & \textbf{100.00} & \textbf{100.00} & \textbf{100.00}  & 18.91 $\pm$ 0.75 & \textit{6.00 $\pm$ 1.5}6  & \textit{3.59 $\pm$ 0.45} & 1.018s \\
\midrule

\multicolumn{8}{c}{\textbf{GMSC} \cite{KaggleGiveMeSomeCredit}} \\
\midrule
Method &
Validity (\%) &
Actionability (\%) &
Causality (\%) &
Plausibility (NLL)$\downarrow$ &
Similarity$\downarrow$ &
Sparsity$\downarrow$ &
Time (s)$\downarrow$ \\
\midrule
VAE            & 0.00 & 0.00 & -- & -- & -- & -- & 0.700s \\
CVAE           & 0.81 & 0.61 & -- & 34.19 $\pm$ 4.81 & 16.83 $\pm$ 1.65 & 7.67 $\pm$ 0.58 & 0.470s \\
DiCE           & \textbf{100.00} & \textit{91.70} & -- & 31.79 $\pm$ 1.99 & 23.85 $\pm$ 3.36 & 6.76 $\pm$ 0.46 & 27.55s \\
FACE(knn)      & 8.91 & 4.26 & -- & 33.44 $\pm$ 2.66 & 17.66 $\pm$ 1.53 & 8.18 $\pm$ 0.56 & 6.680s \\
FACE($\epsilon$)& 6.88 & 2.84 & -- & 33.18 $\pm$ 1.94 & 18.55 $\pm$ 2.83 & 8.11 $\pm$ 0.43 & 9.250s \\
PROPLACE       & 48.35 & 37.20 & -- & 33.48 $\pm$ 1.14 & 15.87 $\pm$ 0.55 & 7.40 $\pm$ 0.21 & \textit{0.250s} \\
LiCE  & 6.05 & 4.43 & -- & \textit{27.41} $\pm$ 2.49 & \textit{15.21 $\pm$ 3.82} & \textit{5.20 $\pm$ 0.55} & 122.50s \\
\midrule\midrule
\textbf{PAR}
& 85.70
& \textbf{100.00}
& --
& 16.96 $\pm$ 3.83
& 11.86 $\pm$ 3.05
& 4.65 $\pm$ 0.20
& \textbf{0.209s} \\
\textbf{PAR (+LS)}
& 99.21
& \textbf{100.00}
& --
& \textbf{15.13 $\pm$ 2.72}
&\textbf{ 6.28 $\pm$ 0.81}
& \textbf{2.22 $\pm$ 0.50}
& 1.261s \\
\bottomrule
\end{tabular}%
}
\end{table*}

\section{Experiments and Results}
\label{sec:exp_res}

We evaluate PAR on three tabular benchmarks: Adult \cite{BeckerKohavi1996Adult}, German Credit \cite{Hofmann1994GermanCredit}, and GMSC \cite{KaggleGiveMeSomeCredit}, following the splits used in LiCE \cite{Mothilal2020DiCE}. We used the DeeProb-Kit package \cite{loconte2022deeprob} to define and train SPNs, which implements structure learning and parameter estimation based on the LearnSPN algorithm \cite{Poon2011SPN}. A candidate counterfactual is considered valid if it induces a strict decision flip under the fold-specific classifier at a fixed decision threshold (For details refer Appendix \ref{app:gmsc-tau}, \cite{Youden1950Index}). Plausibility is reported as negative log-likelihood (NLL) under the positive-class generative model $p^+$. Actionability is evaluated using feature-level constraints (immutability and monotonicity). In addition, we report a diagnostic causality metric based on explicit increase-implies-increase rules between features. Similarity is measured using the $\ell_1$ distance weighted by inverse Median Absolute Deviation (MAD), following prior work \cite{Wachter2017CounterfactualSSRN}. Sparsity is measured as the number of mutable features whose values differ between the factual instance and the recourse. Runtime is also computed as the median time required to generate a single counterfactual. To ensure a strict and fair comparison, we adopt the exact fold structure and split files released with LiCE for all methods. Denied factuals are identified using the fold-specific classifier. Baseline methods are evaluated using their native raw-space representations and classifiers. We design our experiments to answer the following questions.

\paragraph{Q1: How does PAR compare to state-of-the-art methods?}
Based on Table \ref{tab:recourse_main}, across the Adult, Credit, and GMSC datasets, PAR exhibits a strong and consistent performance profile when compared with a wide range of existing counterfactual explanation methods. PAR achieves high validity across all datasets and remains competitive with the strongest baselines. While certain optimization-based methods achieve marginally higher validity in some settings, these gains are typically accompanied by significantly higher computational cost. Incorporating local search further improves PAR’s validity, indicating that the learned generator provides strong initial counterfactuals that can be refined when needed. PAR consistently achieves perfect actionability across all datasets, matching or exceeding optimization-based methods such as MIO and LiCE variants, and substantially outperforming generative baselines such as VAE, CVAE, DiCE, and FACE. PAR also achieves low native NLL values, indicating that the generated recourses are highly plausible under its learned data distribution.

A key advantage of PAR is its runtime performance. PAR is among the fastest methods across all datasets, often outperforming optimization-based approaches by more than an order of magnitude. Even with the addition of local search, PAR remains significantly faster than most baselines, underscoring the benefit of amortized generation for scalable recourse.

PAR’s main tradeoff lies in similarity and sparsity. In general, achieving higher plausibility often requires sacrificing proximity, and this tension is evident across methods. PAR explicitly navigates this tradeoff by prioritizing likelihood while inducing only modest changes in similarity and sparsity. While some optimization-based methods achieve slightly sparser or more proximal recourses, these improvements typically come at the cost of substantially higher runtime and, in some cases, reduced plausibility. On GMSC in particular, PAR strikes a favorable balance, achieving better similarity and sparsity than most baselines while simultaneously attaining strong plausibility, as reflected by low NLL.

PAR achieves a substantially higher mean $\hat{y}$ than LiCE, indicating stronger and more decisive decision flips. Although LiCE can outperform PAR on proximity, this advantage does not translate to robustness: under classifier changes, LiCE’s counterfactuals frequently fail, whereas PAR remains stable. We report the robustness analysis in Appendix \ref{app:robustness_yhat}.

\paragraph{Q2: What is the role of the individual losses on the total objective?}
We conduct extensive ablations to analyze the contribution of individual loss terms across all three datasets. Due to space constraints, we focus the discussion on the interaction between plausibility losses and validity, as this relationship is central to the behavior of the proposed objective. The effects of the remaining loss components are reported and analyzed in the Appendix \ref{app:ablations}.

Table \ref{tab:loss_ablation_no_ls} shows that across datasets, plausibility and validity play markedly different roles in inducing effective recourse. On Adult, plausibility optimization exhibits a strong implicit validity effect: even when the explicit validity loss is ablated, enabling both plausibility terms preserves high validity with only a modest drop in the predicted score $\hat{y}$. Enabling only the positive-class plausibility term achieves the lowest NLL while maintaining comparable validity, indicating that movement toward high-density accepted regions is often sufficient to cross the decision boundary in this relatively simple decision landscape. In contrast, Credit shows a strong dependence on explicit validity supervision. Validity optimization alone yields poor validity, and plausibility alone fails almost entirely despite reasonable likelihood, demonstrating that classifier feedback without distributional guidance—and vice versa—is insufficient. Reliable recourse on Credit emerges only when validity is combined with plausibility, particularly the positive-class term, which yields perfect validity; disabling plausibility while retaining validity leads to unstable or degenerate solutions. GMSC lies between these extremes: plausibility alone again fails to induce valid recourse, while the validity loss by itself is partially effective but produces less plausible counterfactuals with higher NLL. As with Credit, the best performance is achieved when validity supervision is paired with the positive-class plausibility term, yielding high validity and the lowest NLL. Across all datasets, removing both plausibility and validity terms leads to degenerate behavior with collapsed validity, highlighting that while plausibility can implicitly induce validity in simpler settings, robust and consistently high-quality recourse in more complex datasets typically requires explicit classifier supervision combined with distributional guidance—most effectively through the positive-class plausibility term.

\paragraph{Q3: What is the impact of the neighborhood encoder?}
We quantify the impact of neighborhood statistics by comparing the recourse generator’s output logits over mutable actions with and without the neighborhood encoder, while conditioning on denied instances that share identical immutable attributes. Across all datasets, removing neighborhood information induces large and systematic shifts in the generator’s action preference distribution. On Adult, between $28\%$ and $59\%$ of mutable-action logits change by more than $0.5$, with relative $\ell_1$ differences ranging from $26\%$ to $91\%$. The effect is substantially stronger on Credit, where over $94\%$ of mutable-action logits exceed the $0.5$ change threshold, with relative $\ell_1$ changes of $33\%$–$42\%$, reflecting the increased reliance on local structure in high-cardinality, constraint-heavy domains. A similar but slightly attenuated pattern is observed on GMSC, where $67\%$–$73\%$ of mutable-action logits change by more than $0.5$, with relative $\ell_1$ differences of approximately $27\%$–$33\%$. Taken together, these results show that neighborhood statistics are not redundant with immutable conditioning or global plausibility, but play a central role in shaping the generator’s action distribution even when immutable attributes are held fixed. See Appendix \ref{app:logit-change}

\paragraph{Q4: What is the impact of local search?}
\begin{figure}[]
    \centering
    \begin{subfigure}[t]{0.32\linewidth}
        \centering
        \includegraphics[width=\linewidth]{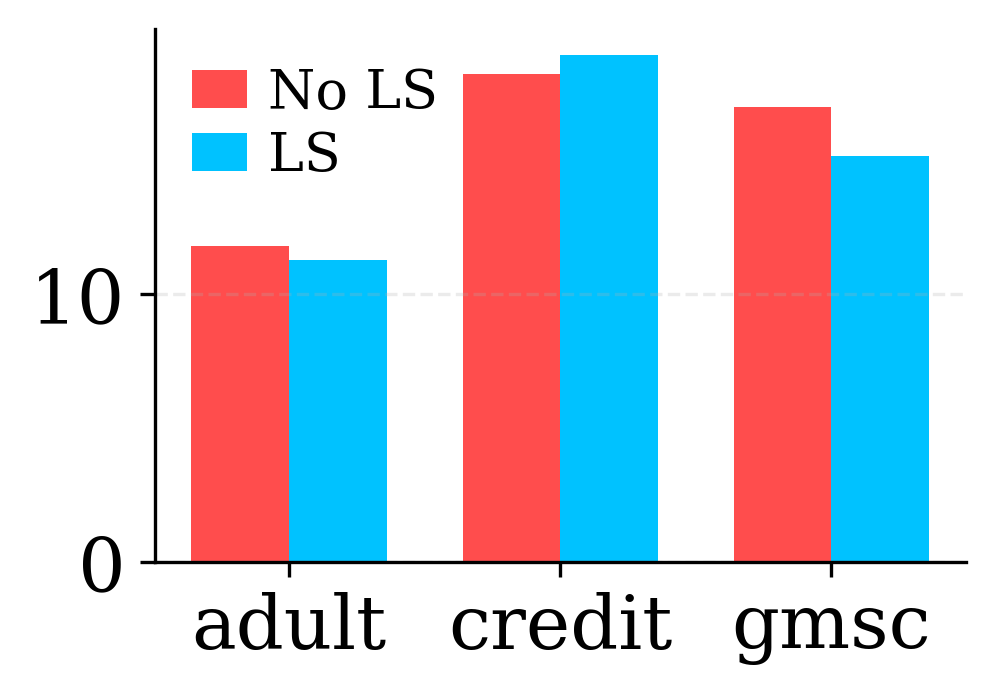}
        \caption{Plausibility (NLL)}
        \label{fig:ls_ablation_nll}
    \end{subfigure}\hfill
    \begin{subfigure}[t]{0.32\linewidth}
        \centering
        \includegraphics[width=\linewidth]{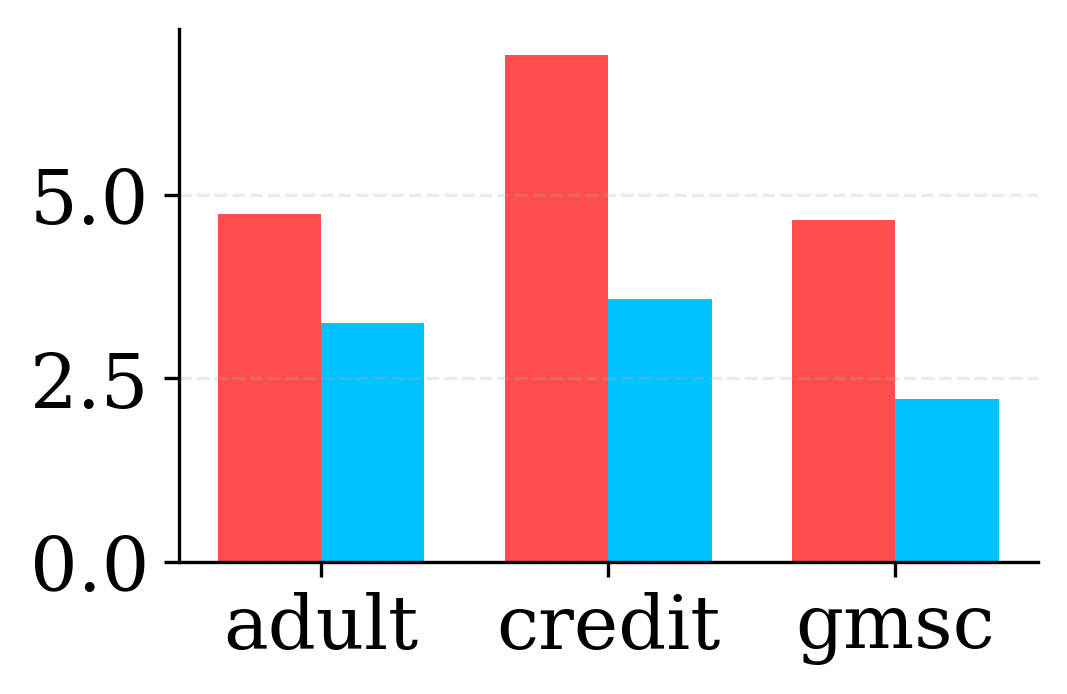}
        \caption{Sparsity}
        \label{fig:ls_ablation_sparsity}
    \end{subfigure}\hfill
    \begin{subfigure}[t]{0.32\linewidth}
        \centering
        \includegraphics[width=\linewidth]{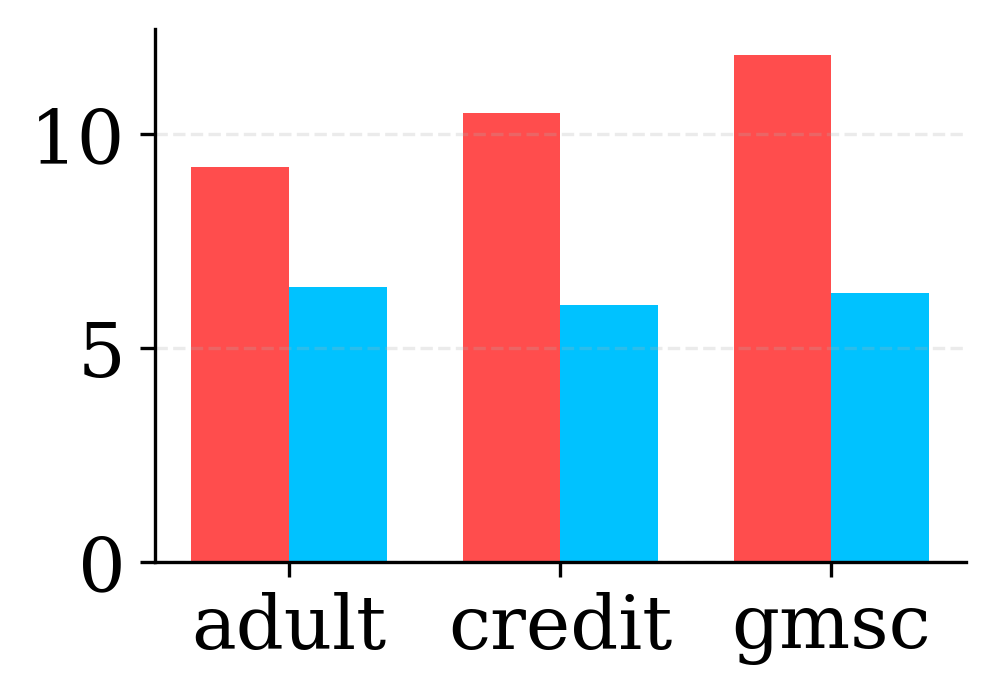}
        \caption{Similarity}
        \label{fig:ls_ablation_similarity}
    \end{subfigure}
    \caption{Effect of local search (LS) on plausibility, sparsity, and similarity across datasets. Each bar pair compares counterfactuals before and after applying LS.}
    \label{fig:ls_ablation_threeplots}
\end{figure}

Figure~\ref{fig:ls_ablation_threeplots} shows that local search  improves proximity and sparsity while preserving constraint satisfaction. Across all datasets, adding local search reduces sparsity substantially and yields markedly lower similarity cost, meaning the selected counterfactuals stay closer to the factual instance. For plausibility, the effect is not uniform; it improves on Adult and GMSC, but slightly worsens on Credit, indicating a mild proximity-plausibility trade-off in that dataset. Finally, there are no causality and monotonicity violations (and hence not included in the Figure) with and without local search, showing that local search does not break the hard constraints enforced in PAR.

\begin{table}[H]
\small
\setlength{\tabcolsep}{5pt}
\renewcommand{\arraystretch}{1.15}
\caption{Loss-term ablation results without local search (mean $\pm$ std over 5 folds).
\xmark\ = ablated; \cmark\ = enabled.}
\label{tab:loss_ablation_no_ls}

\begin{subtable}{\linewidth}
\raggedright
\caption{ADULT}
\label{tab:loss_ablation_adult}
\begin{tabular}{cccccc}
\toprule
\textbf{p+} & \textbf{p-} & \textbf{Val} & \textbf{NLL} & \textbf{$\hat{y}$} & \textbf{validity} \\
\midrule
\xmark & \xmark & \cmark & 18.67 $\pm$ 2.47 & 0.78 $\pm$ 0.08 & 0.89 $\pm$ 0.08 \\
\cmark & \cmark & \xmark & 10.36 $\pm$ 1.53 & 0.75 $\pm$ 0.07 & 0.87 $\pm$ 0.06 \\
\cmark & \xmark & \cmark & 7.91 $\pm$ 0.52 & 0.71 $\pm$ 0.07 & 0.88 $\pm$ 0.07 \\
\xmark & \xmark & \xmark & 13.70 $\pm$ 0.98 & 0.13 $\pm$ 0.02 & 0.06 $\pm$ 0.02 \\
\bottomrule
\end{tabular}
\end{subtable}


\begin{subtable}{\linewidth}
\raggedright
\caption{CREDIT}
\label{tab:loss_ablation_credit}
\begin{tabular}{cccccc}
\toprule
\textbf{p+} & \textbf{p-} & \textbf{Val} & \textbf{NLL} & \textbf{$\hat{y}$} & \textbf{validity} \\
\midrule
\xmark & \xmark & \cmark & 19.12 $\pm$ 0.95 & 0.31 $\pm$ 0.12 & 0.19 $\pm$ 0.13 \\
\cmark & \cmark & \xmark & 18.88 $\pm$ 0.59 & 0.21 $\pm$ 0.07 & 0.05 $\pm$ 0.05 \\
\cmark & \xmark & \cmark & 15.52 $\pm$ 0.45 & 0.95 $\pm$ 0.02 & 1.00 $\pm$ 0.00 \\
\xmark & \xmark & \xmark & 20.32 $\pm$ 0.39 & 0.10 $\pm$ 0.03 & 0.01 $\pm$ 0.02 \\
\bottomrule
\end{tabular}
\end{subtable}

\begin{subtable}{\linewidth}
\raggedright
\caption{GMSC}
\label{tab:loss_ablation_gmsc}
\begin{tabular}{cccccc}
\toprule
\textbf{p+} & \textbf{p-} & \textbf{Val} & \textbf{NLL} & \textbf{$\hat{y}$} & \textbf{validity} \\
\midrule
\xmark & \xmark & \cmark & 17.32 $\pm$ 2.54 & 0.45 $\pm$ 0.27 & 0.71 $\pm$ 0.29 \\
\cmark & \cmark & \xmark & 14.49 $\pm$ 0.58 & 0.04 $\pm$ 0.02 & 0.18 $\pm$ 0.19 \\
\cmark & \xmark & \cmark & 11.30 $\pm$ 0.90 & 0.15 $\pm$ 0.03 & 0.96 $\pm$ 0.062 \\
\xmark & \xmark & \xmark & 16.16 $\pm$ 0.88 & 0.01 $\pm$ 0.00 & 0.04 $\pm$ 0.05 \\
\bottomrule
\end{tabular}
\end{subtable}

\end{table}

\section{Conclusion}
PAR formulates algorithmic recourse as constrained MAP inference over probabilistic circuits, enforcing plausibility directly through the data distribution rather than heuristic regularization. Once the probabilistic circuits and classifier are trained, recourse generation does not require access to the original training data and relies only on samples from the learned models, making PAR robust to class imbalance and suitable for deployment when training data is unavailable. The same generator can be reused as long as the immutable feature set and structural constraints remain fixed, amortizing training cost across users. However, PAR assumes that data constraints are explicitly provided, and changes to the immutable feature set (e.g., user-specific immutability preferences) require retraining the generator. In addition, the local search refinement improves sparsity and constraint satisfaction but adds extra inference-time overhead.

\bibliographystyle{named}
\bibliography{ijcai26}
\onecolumn
\appendix
\input{appendix}
\end{document}

%% file: appendix.tex





\appendix


\section{Data Modifications and Constraints}
\label{app:data_constraints}

All experiments are conducted using the same pruned datasets and the same immutability, monotonicity, and causality constraints as defined in LiCE.
We do not introduce any additional data filtering or constraint modifications.

\subsection{Dataset Pruning and Feature Removal}
\label{app:data_pruning}
This section summarizes the resulting datasets and constraints after applying the LiCE preprocessing pipeline.
\begin{table}[H]
\centering
\small
\setlength{\tabcolsep}{6pt}
\renewcommand{\arraystretch}{1.15}
\caption{Dataset-specific pruning rules adopted from LiCE.}
\label{tab:data_pruning_rules}
\begin{tabular}{p{2.2cm} p{11.2cm}}
\toprule
\textbf{Dataset} & \textbf{Pruning Criteria} \\
\midrule
GMSC &
MonthlyIncome $<50{,}000$;
RevolvingUtilizationOfUnsecuredLines $<1$;
NumberOfTime30--59DaysPastDueNotWorse $<10$;
DebtRatio $<2$;
NumberOfOpenCreditLinesAndLoans $<40$;
NumberOfTimes90DaysLate $<10$;
NumberRealEstateLoansOrLines $<10$;
NumberOfTime60--89DaysPastDueNotWorse $<10$;
NumberOfDependents $<10$ \\
\midrule
Adult &
Samples with missing values removed; features \texttt{fnlwgt}, \texttt{education-num},
\texttt{native-country}, \texttt{capital-gain}, \texttt{capital-loss} removed \\
\midrule
Credit &
No samples or features removed \\
\bottomrule
\end{tabular}
\end{table}

\subsection{Feature Types After Pruning}
\label{app:feature_types}
\begin{table}[H]
\centering
\small
\setlength{\tabcolsep}{6pt}
\renewcommand{\arraystretch}{1.15}
\caption{Resulting datasets after LiCE preprocessing.}
\label{tab:dataset_stats}
\begin{tabular}{lcccc}
\toprule
\textbf{Dataset} & \textbf{Samples} & \textbf{Features} & \textbf{Categorical} & \textbf{Continuous / Discrete} \\
\midrule
GMSC   & 113{,}595 & 10 & 0 & 7 discrete, 3 continuous \\
Adult  & 47{,}876  & 9  & 5 cat, 1 bin, 1 ord & 2 discrete \\
Credit & 1{,}000   & 20 & 10 cat, 2 bin, 1 ord & 7 numeric \\
\bottomrule
\end{tabular}
\end{table}

\subsection{Immutability, Monotonicity, and Causality Constraints}
\label{app:constraints_summary}
\begin{table}[H]
\centering
\small
\setlength{\tabcolsep}{6pt}
\renewcommand{\arraystretch}{1.15}
\caption{Immutability, monotonicity, and causality constraints adopted from LiCE.}
\label{tab:data_constraints}
\begin{tabular}{p{2.1cm} p{4.6cm} p{3.2cm} p{4.6cm}}
\toprule
\textbf{Dataset} & \textbf{Immutable Features} & \textbf{Monotone Constraints} & \textbf{Causal Constraints} \\
\midrule
GMSC &
NumberOfDependents &
Age non-decreasing &
None \\
\midrule
Adult &
Race, Sex &
Age non-decreasing; Education non-decreasing &
Education $\uparrow \Rightarrow$ Age $\uparrow$ \\
\midrule
Credit &
Number of people liable for maintenance; Personal status and sex; Foreign worker &
Age non-decreasing &
Present residence since $\uparrow \Rightarrow$ Age $\uparrow$; \newline
Present employment since $\uparrow \Rightarrow$ Age $\uparrow$ \\
\bottomrule
\end{tabular}
\end{table}

\section{Discretization}
\label{app:preprocessing}

\subsection{Discretization Procedure}
\label{app:discretization}

Each feature is assigned to one of the following discretization types based on its observed data type and cardinality in the training data:
\begin{itemize}
\item \textbf{Ordered categorical features}, for which a fixed semantic ordering is defined,
\item \textbf{Unordered categorical features}, encoded via label encoding without imposing an order,
\item \textbf{Discrete numeric features}, defined as numeric features with low cardinality,
\item \textbf{Continuous numeric features}, discretized using quantile-based binning.
\end{itemize}

For continuous numeric features, quantile-based binning is performed using quantiles computed on the training split only.
If quantile binning degenerates due to limited variability or repeated values, the feature is treated as a discrete numeric feature using its unique observed values.
This fallback guarantees that every feature admits at least one valid discrete category.

The number of discrete categories per feature is fixed at discretizer fit time and validated to be strictly positive.
These per-feature cardinalities define the discrete domain used throughout the pipeline, including classifier training, probabilistic circuit construction, counterfactual generation, and constraint enforcement.

\subsection{Ordered Feature Handling and Safety Guarantees}
\label{app:ordered_features}
For features subject to monotonicity or causality constraints, preserving semantic ordering during discretization is essential. For ordinal categorical features, an explicit, semantically meaningful ordering is specified prior to discretization.
If a feature participates in monotonicity or causality constraints but appears categorical-like without an explicit ordering, numeric coercion is applied.
Numeric coercion converts the feature into a numeric representation that reflects its underlying ordered semantics.
If numeric coercion is unreliable or ambiguous, execution is halted rather than silently imposing an arbitrary ordering.
This design choice ensures that comparisons such as increases and decreases are semantically meaningful.

\section{Bin Coverage and Bin Fidelity}
\label{app:bin_fidelity}

We evaluate whether the discretization map learned from the training split adequately supports the raw test distribution.
To do so, we compute two diagnostics on the raw test features using the discretizer metadata saved at training time (the per-feature domain definition, including category vocabularies, discrete value sets, and binning type).

\subsection{In-domain indicator}
\label{app:in_domain_indicator}

Let $X \in \mathbb{R}^{N \times D}$ denote the raw test feature matrix, and let $\mathcal{D}_j$ denote the training-time domain induced by the discretizer for feature $j$.
For each test instance $i$ and feature $j$, we define an in-domain indicator
\[
\mathbb{I}_{ij} =
\begin{cases}
1, & \text{if } x_{ij} \in \mathcal{D}_j,\\
0, & \text{otherwise.}
\end{cases}
\]

The domain membership rule depends on the discretization type of feature $j$:
\begin{itemize}
    \item \textbf{Categorical (ordered or unordered).} $\mathbb{I}_{ij}=1$ iff the raw category is in the training vocabulary and is not mapped to the reserved missing/unknown token.
    \item \textbf{Discrete numeric.} $\mathbb{I}_{ij}=1$ iff the value is finite and exactly equals one of the discrete values observed in training.
    \item \textbf{Binned numeric.} $\mathbb{I}_{ij}=1$ iff the value is finite (since quantile binning defines bins over the numeric support).
\end{itemize}
Importantly, these checks are performed \emph{without} counting fallback behaviors (e.g., mapping unknown categories to an NA token or snapping out-of-set discrete values to the nearest training value) as ``in-domain''; those are treated as out-of-domain.

\subsection{Bin coverage and bin fidelity}
\label{app:coverage_fidelity_defs}

Using $\mathbb{I}_{ij}$, we define:
\begin{align}
\textbf{Bin Coverage} &:= \frac{1}{ND}\sum_{i=1}^{N}\sum_{j=1}^{D}\mathbb{I}_{ij},\\
\textbf{Bin Fidelity} &:= \frac{1}{N}\sum_{i=1}^{N}\prod_{j=1}^{D}\mathbb{I}_{ij}.
\end{align}
Bin Coverage measures the fraction of individual feature values in the raw test set that lie within the discretizer-defined domain.
Bin Fidelity is stricter: it measures the fraction of test instances for which \emph{all} features are in-domain simultaneously.

Table~\ref{tab:bin_fidelity} reports Bin Coverage and Bin Fidelity for each dataset (aggregated over folds).
Both metrics are effectively $1.0$, indicating that the training-time discretization domains cover the raw test distributions with negligible mismatch.

\begin{table}[t]
\centering
\small
\setlength{\tabcolsep}{6pt}
\renewcommand{\arraystretch}{1.15}
\caption{Bin coverage and bin fidelity on the raw test splits under the discretizer-defined feature domains. Higher is better; both metrics are effectively $1.0$, indicating negligible domain mismatch.}
\label{tab:bin_fidelity}
\begin{tabular}{lcc}
\toprule
\textbf{Dataset} & \textbf{Bin Coverage} & \textbf{Bin Fidelity} \\
\midrule
Adult  & $1.0000$   & $1.0000$ \\
Credit & $1.0000$   & $1.0000$ \\
GMSC   & $0.999996$ & $0.999956$ \\
\bottomrule
\end{tabular}
\end{table}

\section{Model Training and Hyperparameter Selection}
\label{app:model_setup}

\subsection{Classifier Training}
\label{sec:classifier-training}

For each dataset and fold, we train a binary classifier using the fold-specific training split provided by the experimental data layout.
Specifically, the classifier is trained on the discretized representation of $(X_{\text{train}}, y_{\text{train}})$ loaded from the fold directory, and it is subsequently used to (i) identify denied instances requiring recourse and (ii) evaluate counterfactual validity during generation and post-processing.

\paragraph{Input representation.}
Let $x \in \{0,\dots,C_1\!-\!1\}\times\cdots\times\{0,\dots,C_D\!-\!1\}$ denote a discretized instance, where feature $j$ has $C_j$ categories.
We convert $x$ into a single concatenated one-hot vector of dimension $\sum_{j=1}^{D} C_j$ by one-hot encoding each feature and concatenating the resulting blocks.
This representation is used consistently for classifier training and for evaluating counterfactuals.

\paragraph{Model architecture.}
We use a fully connected neural network with two hidden layers:
\[
\text{Linear}(d,20) \rightarrow \text{ReLU} \rightarrow
\text{Linear}(20,10) \rightarrow \text{ReLU} \rightarrow
\text{Linear}(10,1),
\]
followed by a sigmoid to obtain a predicted acceptance score $\hat{y}(x)\in[0,1]$.

\paragraph{Training procedure.}
The model is trained using binary cross-entropy loss and the Adam optimizer with learning rate $10^{-3}$, batch size $64$, and a fixed number of epochs.

The classifier produces a continuous score, $\hat{y}(x)$ which is thresholded to obtain a binary decision.
A counterfactual is considered valid if $\hat{y}(x^{\mathrm{cf}})\ge \tau$, and denied factual instances are those with $\hat{y}(x_f)<\tau$.
We use a dataset-specific threshold $\tau$, including a tuned operating point for GMSC as described in Section~\ref{app:gmsc-tau}.

\subsection{Threshold Selection for GMSC}
\label{app:gmsc-tau}

The GMSC dataset exhibits a highly skewed score distribution under the trained classifier: predicted acceptance probabilities for factual instances are concentrated near zero, with mean values well below $0.1$.
As a result, using a fixed decision threshold such as $\tau=0.5$ would classify almost all instances as negative, leading to an uninformative definition of denied instances and counterfactual validity.
We therefore determine the decision threshold for GMSC using a data-driven procedure.

\paragraph{Internal validation split.}
For each fold, we construct an internal validation set by randomly partitioning the provided training split into $80\%$ training and $20\%$ validation.
The discretizer and classifier are fitted using only the internal training portion and then applied unchanged to the validation portion.

\paragraph{Classifier training.}
We train a single neural network classifier using binary cross-entropy loss and the Adam optimizer.
No model averaging or repeated training is performed; results correspond to one trained model per fold.

\paragraph{Threshold sweep and selection.}
Let $\hat{y}(x)\in[0,1]$ denote the predicted acceptance probability on the validation set.
We evaluate a grid of candidate thresholds $\tau \in \{0.01,0.02,\dots,0.99\}$ and compute, for each $\tau$, the true positive rate $\mathrm{TPR}(\tau)$ and false positive rate $\mathrm{FPR}(\tau)$.
We select the operating threshold $\tau^\star$ by maximizing Youden’s $J$ statistic,
\[
\tau^\star \;=\; \arg\max_{\tau}\big(\mathrm{TPR}(\tau) - \mathrm{FPR}(\tau)\big),
\]
which corresponds to the threshold that maximizes balanced separation between positive and negative classes.
Mean $\tau$ across all folds  is $0.06$ which we used for gmsc as the threshold.

\subsection{Hyperparameter Tuning}
\label{sec:hp-tuning}

Hyperparameters are selected using a fold-wise evaluation procedure on the predefined data folds for each dataset. For each candidate configuration, the full recourse pipeline is executed independently on each fold, and performance is aggregated across folds to assess stability and robustness. Configurations that exhibit poor intermediate performance are pruned early to improve efficiency, and the final hyperparameter setting is chosen based on aggregate fold-level metrics.

\subsection{Model Architectures}
\label{app:model_architectures}

\paragraph{Neighborhood Encoder}
The neighborhood encoder summarizes local structure around a denied factual instance using a fixed number of neighbors. For each neighbor, a low-dimensional feature vector is constructed consisting of the neighbor distance and its log-likelihood under the positive-class distribution. Each neighbor feature is passed independently through a two-layer multilayer perceptron (MLP) with ReLU activations. The resulting embeddings are aggregated using mean pooling across neighbors, followed by a linear projection to produce a fixed-dimensional neighborhood representation. This embedding is used to condition the recourse generation process.

\paragraph{Recourse Generator}
The recourse generator is a feedforward neural network that produces counterfactual logits for all mutable features simultaneously. Its input consists of the concatenation of (i) a one-hot encoding of immutable features, (ii) a scalar likelihood-based summary of the factual instance, and (iii) the neighborhood embedding produced by the encoder. The network comprises two fully connected hidden layers with ReLU activations and outputs a single logit vector spanning the discretized categories of all mutable features. These logits are later decoded into a counterfactual instance under the imposed constraints.

\paragraph{Probabilistic Circuit Architecture}
Each feature is modeled using a categorical distribution at the leaf level, with category domains determined by the discretization cardinalities. The probabilistic circuit structure is learned using the \textsc{LearnSPN} algorithm, with leaf parameters estimated via maximum likelihood estimation (MLE). Structure learning is controlled using a minimum row slice size of 200 and a minimum column slice size of 3, and is performed independently for each class using identical hyperparameters.

\section{Logit Change Analysis}
\label{app:logit-change}

To quantify the effect of the neighborhood encoder on the recourse generator, we measure how the generator’s output logits change when the encoder is included versus removed, while keeping all other components fixed.

Let $\ell^{\text{enc}} \in \mathbb{R}^{\sum_j C_j}$ denote the generator logits produced with the neighborhood encoder, and $\ell^{\text{no-enc}}$ the logits produced without it, where the logits are concatenated over all mutable feature categories. We define the logit change using a relative $\ell_1$ metric,
\[
\Delta_{\ell_1} \;=\;
\frac{1}{B} \sum_{i=1}^{B}
\frac{\left\lVert \ell^{\text{enc}}_i - \ell^{\text{no-enc}}_i \right\rVert_1}
{\left\lVert \ell^{\text{no-enc}}_i \right\rVert_1 + \varepsilon},
\]
where $B$ is the number of denied factuals and $\varepsilon$ is a small constant for numerical stability.

This metric captures the magnitude of change induced by the neighborhood encoder relative to the baseline generator output, aggregated over all mutable feature blocks. In addition to the global score, we also report blockwise mean absolute logit changes per mutable feature to localize which features are most affected by the neighborhood context.

\section{Tractable Gradient Computation on Probabilistic Circuit}
\label{app:pc_grad}
Let $\mathcal{C}$ be a smooth and decomposable PC $\mathcal{C}$ defined over discrete variables $X=(X_1, ...,X_D)$. Each variable $X_j$ takes values in $\{0, ..., C_j-1\}$. We consider soft categorical inputs $q = (q_1,...,q_D)$, where each $q_j \in \Delta^{C_j-1}$ is a categorical probability vector. Given a soft input $q_j$, we evaluate the leaf by the multilinear form
\begin{equation}
    l_j(q_j) = \sum_{c=0}^{C_j-1} q_{j,c}\theta_{j,c}.
\end{equation}
All other internal nodes follow the standard PC computation: sum nodes compute convex combiantion of children and product nodes compute products over disjoint scopes. Under this construction, the circuit output $v_\mathcal{C}(q)$ is a multilinear polynomial in the soft inputs $q$. For any feature $j$ and category $c$, the partial derivative $\partial \log v_\mathcal{C}/\partial q_{j,c}$ can be computed by a single bottom-up value computation on $\mathcal{C}$ with a modified input $q_{j\leftarrow e_c}$. Moreover, the full gradient $\nabla_q \log v_\mathcal{C}(q)$ can be computed in time and space linear in the circuit size times the total number of categories.
\par Define $e_c$ as the one-hot vector with $(e_c)_c = 1$. Let $q_{j\leftarrow e_c}$ denote the same soft input as $q$ except that the block $q_j$ is replaced by $e_c$. i.e.,
\begin{equation}
    q_{j \leftarrow e_c} = [q_1;...;q_{j-1};e_c;q_{j+1};...;q_D]
\end{equation}
Then, by multilinearity of $v_\mathcal{C}(q)$ in each block $q_j$, we obtain
\begin{equation}
   v_{\mathcal{C}}(q)=\sum_{c'=0}^{C_j-1} q_{j,c'}\, v_{\mathcal{C}}(q_{j\leftarrow e_{c'}}).
\end{equation}

Differentiating w.r.t $q_{j,c}$ yields
\begin{equation}
    \frac{\partial v_\mathcal{C}}{\partial q_{j,c}}=v_\mathcal{C}(q_{j \leftarrow e_c})
\end{equation}
\par \textbf{Proposition:} Let $\mathcal{C}$ be a smooth and decomposable PC with categorical leaves. then the gradient of the loss $-\log v_\mathcal{C}(q)$ w.r.t $q$ can be computed exactly using bottom-up computations on $\mathcal{C}$. For each feature $j$, computing all partial derivatives $\{\partial \log v_\mathcal{C}(q)/\partial q_{j,c}\}_{c=1}^{C_j}$ requires $C_j$ circuit evaluations, each linear in the size of $\mathcal{C}$, where $C_j$ denotes the number of discrete categories of feature $j$.

\par \textbf{Proof:} w.l.o.g, assume the root node $r$ of $\mathcal{C}$ is a sum node and that the circuit alternates between sum and product nodes. Consider the partial derivative w.r.t a component $q_{j,c}$.

Because the circuit is smooth, all children of any sum node with scope $X_j$ also has scope $X_j$. Thus for a sum node $n$,
\begin{equation}
    \frac{\partial v(n,q)}{\partial q_{j,c}} = \sum_{m \in ch(n)} w_{n,m}\frac{\partial v(m, q)}{\partial q_{j,c}}
\end{equation}
by the sum rule of differentiation.
Now consider a product node $n$. By decomposability, at most one child $m^* \in ch(n)$ has $X_j$ in its scope. All other children are independent of $q_{j,c}$ and can be treated as constants. Hence,
\begin{equation}
\frac{\partial v(n, q)}{\partial q_{j,c}}
=
\frac{\partial v(m^*;q)}{\partial q_{j,c}}
\prod_{m\in ch(n)\setminus\{m^*\}} v(m,q).
\end{equation}

At a categorical leaf $l_j$ over features $X_j$, we have
\begin{equation}
    v(l_j,q) = \sum_{c'=0}^{C_j-1} q_{j,c'}\theta_{j,c'}.
\end{equation}
therefore
\begin{equation}
    \frac{\partial v(l_j,q)}{\partial q_{j,c}} = \theta_{j,c}
\end{equation}
Combining the above observations, the partial derivative propagates through the circuit along exactly one branch at each product node and linearly through sum nodes.

\section{Hard Constraint Enforcement}
\label{app:constraints}

\subsection{Monotonicity Masking}
\label{app:monotonicity_masking}
\par \textbf{Hard monotonicity constraint.}
Let $x^- \in \{0,\dots,C_j-1\}^D$ denote the factual instance in discretized (bin) space, and let
$l_j \in \mathbb{R}^{C_j}$ be the generator logits for feature $j$.
For each monotone feature $j \in \mathcal{M}^\uparrow$, only non-decreasing changes are allowed:
\[
x^+_j \ge x^-_j .
\]
We enforce this deterministically by masking all illegal logits,
\[
\tilde l_{j,a} =
\begin{cases}
l_{j,a}, & a \ge x^-_j,\\
M, & a < x^-_j,
\end{cases}
\qquad, 
q_j(a) = \mathrm{softmax}(\tilde l_j)_a .
\]where $M$ is a large integer.
As a result, any decoded counterfactual automatically satisfies monotonicity.

\subsection{Causal Masking}
\label{app:causal_masking}
Causal relations are specified as ordered pairs $(e,c)$, meaning that an increase in the effect
feature $e$ is only allowed if the cause feature $c$ also increases:
\[
(x^+_e > x^-_e) \;\Rightarrow\; (x^+_c > x^-_c).
\]
We enforce this constraint both during training (via constrained soft assignments) and at
decoding time (via feasible discrete selection).

\emph{Case 1: both $e$ and $c$ are mutable.}
Let $\tilde l_e \in \mathbb{R}^{C_e}$ and $\tilde l_c \in \mathbb{R}^{C_c}$ be the (already
monotone-masked, if applicable) logits.
We construct joint logits
\[
L^{(e,c)}_{a_c,a_e} = \tilde l_{c,a_c} + \tilde l_{e,a_e},
\]
and mask all illegal assignments,
\[
\tilde L^{(e,c)}_{a_c,a_e} =
\begin{cases}
L^{(e,c)}_{a_c,a_e}, &
(a_e \le x^-_e) \;\vee\; (a_c > x^-_c),\\
-\infty, & \text{otherwise}.
\end{cases}
\]
A valid joint distribution is obtained by applying a softmax over all $(a_c,a_e)$ pairs.
The resulting marginals
$q_c(a_c) = \sum_{a_e} q_{c,e}(a_c,a_e)$ and
$q_e(a_e) = \sum_{a_c} q_{c,e}(a_c,a_e)$
are used as the per-feature soft assignments during training.
At decoding time, feasibility is guaranteed by joint argmax:
\[
(x^+_c, x^+_e) = \arg\max_{a_c,a_e} \tilde L^{(e,c)}_{a_c,a_e}.
\]

\emph{Case 2: $e$ mutable, $c$ immutable.}
Since the cause cannot increase, the rule implies that the effect must not increase.
We therefore clamp the effect logits as
\[
\tilde l_{e,a} \leftarrow -\infty \quad \text{for all } a > x^-_e,
\]
and decode $x^+_e = \arg\max_a \tilde l_{e,a}$.

\emph{Case 3: $e$ immutable.}
The constraint is trivially satisfied because $x^+_e = x^-_e$.

\section{Local Search and Post-Processing}
\label{app:local_search}

\begin{algorithm}[H]
\caption{\textsc{RepairCausality}: Minimal Group Repair}
\label{alg:repair}
\small
\begin{algorithmic}[1]
\Require candidate $c$, factual $x_f$, grouped rules $(u \mapsto \mathcal{E}(u))$
\Ensure repaired candidate $c$

\For{each cause $u$}
    \If{$c_u \le (x_f)_u$} \Comment{cause is not strictly increased}
        \For{each effect $v \in \mathcal{E}(u)$}
            \If{$c_v > (x_f)_v$} \State $c_v \gets (x_f)_v$ \EndIf \Comment{clamp effect down}
        \EndFor
    \EndIf
\EndFor
\State \Return $c$
\end{algorithmic}
\end{algorithm}

\begin{algorithm}[H]
\caption{\textsc{Sparsify}: Validity-Preserving Feature Reversion}
\label{alg:sparsify}
\small
\begin{algorithmic}[1]
\Require valid $c$, factual $x_f$, mutable set $\mathcal{M}$, budget $B$
\Require constraints $(\mathcal{I},\mathcal{K},u\mapsto\mathcal{E}(u))$, classifier $h$, threshold $\tau$
\Require optional $\log p_+(\cdot)$ and $\Delta_{\max}$
\Ensure sparsified $c$

\State $\ell \gets \log p_+(c)$ if available
\Repeat
    \State $\text{changed} \gets \textbf{false}$
    \State $\mathcal{S} \gets \{j\in\mathcal{M}: c_j \neq (x_f)_j\}$ sorted by $|c_j-(x_f)_j|$
    \For{each $j \in \mathcal{S}$}
        \State $\tilde{c} \gets c$; $\tilde{c}_j \gets (x_f)_j$
        \State $\tilde{c} \gets \textsc{RepairCausality}(\tilde{c},x_f,u\mapsto\mathcal{E}(u))$
        \If{\textbf{not} \textsc{WithinBudget}$(\tilde{c},x_f,\mathcal{M},B)$ \textbf{or} \textbf{not} \textsc{Feasible}$(\tilde{c},x_f,\mathcal{I},\mathcal{K},u\mapsto\mathcal{E}(u))$}
            \State \textbf{continue}
        \EndIf
        \If{$h(\textsc{OneHot}(\tilde{c})) < \tau$} \State \textbf{continue} \EndIf
        \If{$\log p_+$ available \textbf{and} $\Delta_{\max}$ set}
            \If{$\log p_+(\tilde{c}) < \ell - \Delta_{\max}$} \State \textbf{continue} \EndIf
            \State $\ell \gets \log p_+(\tilde{c})$
        \EndIf
        \State $c \gets \tilde{c}$; $\text{changed} \gets \textbf{true}$
        \State \textbf{break}
    \EndFor
\Until{\textbf{not} $\text{changed}$}
\State \Return $c$
\end{algorithmic}
\end{algorithm}

\begin{algorithm}[H]
\caption{Strict Two-Phase Local Search (PAR, discretized space)}
\label{alg:ls_main}
\small
\begin{algorithmic}[1]
\Require factual $x_f$, initial candidate $c^{(0)}$
\Require mutable $\mathcal{M}$, immutable $\mathcal{I}$, monotone $\mathcal{K}$
\Require cardinalities $\{C_j\}_{j=1}^p$
\Require classifier $h(\cdot)$ and threshold $\tau$
\Require grouped causal rules $(u \mapsto \mathcal{E}(u))$; optional $\log p_+(\cdot)$ and $\Delta_{\max}$
\Ensure refined counterfactual $c^\star$

\State $B \gets \textsc{Ham}_{\mathcal{M}}(c^{(0)},x_f)$
\State $c \gets c^{(0)}$; \For{$j\in\mathcal{I}$} $c_j \gets (x_f)_j$ \EndFor
\State $c \gets \textsc{RepairCausality}(c, x_f, u\mapsto \mathcal{E}(u))$

\If{\textbf{not} \textsc{WithinBudget}$(c,x_f,\mathcal{M},B)$ \textbf{or} \textbf{not} \textsc{Feasible}$(c,x_f,\mathcal{I},\mathcal{K},u\mapsto\mathcal{E}(u))$}
    \State $c \gets x_f$
\EndIf

\State $y \gets h(\textsc{OneHot}(c))$
\If{$y \ge \tau$}
    \State \Return \textsc{Sparsify}$(c, x_f, \mathcal{M}, B, \mathcal{I}, \mathcal{K}, u\mapsto\mathcal{E}(u), h, \tau, \log p_+, \Delta_{\max})$
\EndIf

\State $c_{\text{best}} \gets c$; $y_{\text{best}} \gets y$
\State $c_{\text{valid}} \gets \emptyset$

\For{each $j \in \mathcal{M}$}
    \For{$v = 0$ \textbf{to} $C_j-1$}
        \If{$v = c_j$} \State \textbf{continue} \EndIf
        \State $\tilde{c} \gets c$; $\tilde{c}_j \gets v$
        \State $\tilde{c} \gets \textsc{RepairCausality}(\tilde{c}, x_f, u\mapsto \mathcal{E}(u))$
        \If{\textbf{not} \textsc{WithinBudget}$(\tilde{c},x_f,\mathcal{M},B)$ \textbf{or} \textbf{not} \textsc{Feasible}$(\tilde{c},x_f,\mathcal{I},\mathcal{K},u\mapsto\mathcal{E}(u))$}
            \State \textbf{continue}
        \EndIf
        \State $\tilde{y} \gets h(\textsc{OneHot}(\tilde{c}))$
        \State $\tilde{\ell} \gets \log p_+(\tilde{c})$ if available else $0$

        \If{$\tilde{y} \ge \tau$}
            \State update $c_{\text{valid}}$ by key $\big(-\tilde{y},\ \textsc{Ham}_{\mathcal{M}}(\tilde{c},x_f),\ -\tilde{\ell}\big)$
        \Else
            \State update $(c_{\text{best}},y_{\text{best}})$ if $\tilde{y}$ improves (tie-break by $\tilde{\ell}$)
        \EndIf
    \EndFor
\EndFor

\If{$c_{\text{valid}} = \emptyset$}
    \State \Return $c_{\text{best}}$
\EndIf
\State \Return \textsc{Sparsify}$(c_{\text{valid}}, x_f, \mathcal{M}, B, \mathcal{I}, \mathcal{K}, u\mapsto\mathcal{E}(u), h, \tau, \log p_+, \Delta_{\max})$
\end{algorithmic}
\end{algorithm}

\newpage
\section{Additional Experimental Results}
\label{app:additional_results}

\subsection{Loss-Term Ablation Studies}
\label{app:ablations}
All ablation variants are trained independently. Minor deviations from the main results are attributable to stochastic training effects.

\begin{table}[H]
\small
\setlength{\tabcolsep}{5pt}
\renewcommand{\arraystretch}{1.15}
\caption{Loss-term ablation results without local search(mean $\pm$ std over folds).
\cmark\ = ablated; \xmark\ = enabled.}
\label{tab:loss_ablation_all_no_ls}

\begin{subtable}{\textwidth}
\raggedright
\caption{ADULT}
\label{tab:loss_ablation_adult_no_ls}
\begin{tabular}{cccccccccccc}
\toprule
\textbf{\rot{Proximity}} &
\textbf{\rot{Plaus\_p+}} &
\textbf{\rot{Plaus\_p-}} &
\textbf{\rot{Sparsity}} &
\textbf{\rot{Validity}} &
\textbf{\rot{Entropy}} &
\textbf{\rot{PPT\_block}} &
\textbf{NLL} &
\textbf{ $\hat{y}$} &
\textbf{validity} &
\textbf{similarity} &
\textbf{sparsity} \\
\midrule
\cmark & \cmark & \cmark & \cmark & \cmark & \cmark & \cmark & 12.110 ± 2.248 & 0.867 ± 0.023 & 0.969 ± 0.015 & 10.282 ± 1.007 & 5.131 ± 0.412 \\
\xmark & \cmark & \cmark & \cmark & \cmark & \cmark & \cmark & 10.586 ± 0.812 & 0.893 ± 0.036 & 0.960 ± 0.044 & 11.219 ± 0.702 & 5.435 ± 0.321 \\
\cmark & \xmark & \xmark & \cmark & \cmark & \cmark & \cmark & 18.670 ± 2.473 & 0.783 ± 0.082 & 0.896 ± 0.079 & 7.162 ± 0.835 & 3.730 ± 0.274 \\
\xmark & \cmark & \xmark & \cmark & \cmark & \cmark & \cmark & 7.899 ± 0.505 & 0.720 ± 0.067 & 0.882 ± 0.071 & 7.998 ± 0.824 & 4.223 ± 0.257 \\
\xmark & \xmark & \cmark & \cmark & \cmark & \cmark & \cmark & 42.445 ± 0.745 & 0.770 ± 0.057 & 0.899 ± 0.031 & 22.590 ± 1.275 & 6.777 ± 0.093 \\
\cmark & \cmark & \cmark & \xmark & \cmark & \cmark & \cmark & 12.841 ± 3.067 & 0.923 ± 0.028 & 0.974 ± 0.030 & 12.862 ± 1.326 & 5.631 ± 0.491 \\
\cmark & \cmark & \cmark & \cmark & \xmark & \cmark & \cmark & 10.362 ± 1.530 & 0.754 ± 0.070 & 0.869 ± 0.057 & 9.848 ± 1.360 & 4.978 ± 0.503 \\
\cmark & \cmark & \cmark & \cmark & \cmark & \xmark & \cmark & 10.979 ± 0.772 & 0.768 ± 0.024 & 0.990 ± 0.017 & 7.889 ± 1.211 & 3.792 ± 0.427 \\
\cmark & \cmark & \cmark & \cmark & \cmark & \xmark & \cmark & 10.938 ± 1.062 & 0.760 ± 0.028 & 0.990 ± 0.017 & 8.010 ± 1.220 & 3.820 ± 0.438 \\
\cmark & \cmark & \cmark & \xmark & \cmark & \cmark & \cmark & 11.404 ± 1.380 & 0.767 ± 0.056 & 0.995 ± 0.007 & 8.260 ± 1.908 & 3.474 ± 0.677 \\
\cmark & \cmark & \xmark & \cmark & \cmark & \cmark & \cmark & 8.595 ± 0.640 & 0.684 ± 0.033 & 0.964 ± 0.026 & 6.581 ± 0.836 & 3.492 ± 0.248 \\
\cmark & \cmark & \cmark & \cmark & \cmark & \cmark & \xmark & 17.339 ± 2.432 & 0.672 ± 0.012 & 0.939 ± 0.051 & 5.581 ± 0.742 & 2.611 ± 0.191 \\
\cmark & \cmark & \cmark & \cmark & \cmark & \xmark & \xmark & 17.593 ± 1.585 & 0.662 ± 0.039 & 0.925 ± 0.064 & 5.552 ± 0.651 & 2.583 ± 0.159 \\
\cmark & \xmark & \xmark & \cmark & \xmark & \cmark & \cmark & 10.607 ± 0.482 & 0.894 ± 0.037 & 0.965 ± 0.034 & 11.251 ± 0.744 & 5.436 ± 0.347 \\
\cmark & \xmark & \cmark & \cmark & \xmark & \cmark & \cmark & 7.870 ± 0.527 & 0.722 ± 0.072 & 0.883 ± 0.083 & 8.078 ± 0.858 & 4.238 ± 0.277 \\
\cmark & \cmark & \xmark & \cmark & \xmark & \cmark & \cmark & 42.450 ± 0.779 & 0.768 ± 0.056 & 0.899 ± 0.031 & 22.612 ± 1.325 & 6.772 ± 0.107 \\
\cmark & \xmark & \cmark & \cmark & \cmark & \cmark & \cmark & 8.071 ± 0.506 & 0.614 ± 0.047 & 0.786 ± 0.069 & 6.668 ± 0.688 & 3.901 ± 0.254 \\

\bottomrule
\end{tabular}
\end{subtable}
\end{table}

\begin{table}[H]\ContinuedFloat
\small
\setlength{\tabcolsep}{5pt}
\renewcommand{\arraystretch}{1.15}
\begin{subtable}{\textwidth}
\raggedright
\caption{CREDIT}
\label{tab:loss_ablation_credit_no_ls}
\begin{tabular}{cccccccccccc}
\toprule
\textbf{\rot{Proximity}} &
\textbf{\rot{Plaus\_p+}} &
\textbf{\rot{Plaus\_p-}} &
\textbf{\rot{Sparsity}} &
\textbf{\rot{Validity}} &
\textbf{\rot{Entropy}} &
\textbf{\rot{PPT\_block}} &
\textbf{NLL} &
\textbf{$\hat{y}$} &
\textbf{validity} &
\textbf{similarity} &
\textbf{sparsity} \\
\midrule
\cmark & \cmark & \cmark & \cmark & \cmark & \cmark & \cmark & 18.275 ± 1.452 & 0.797 ± 0.090 & 0.929 ± 0.079 & 10.042 ± 1.483 & 6.743 ± 0.540 \\
\xmark & \cmark & \cmark & \cmark & \cmark & \cmark & \cmark & 23.600 ± 2.334 & 0.999 ± 0.001 & 1.000 ± 0.000 & 22.175 ± 3.063 & 11.974 ± 1.101 \\
\cmark & \xmark & \xmark & \cmark & \cmark & \cmark & \cmark & 19.117 ± 0.953 & 0.313 ± 0.119 & 0.198 ± 0.130 & 9.535 ± 0.826 & 6.356 ± 0.422 \\
\xmark & \cmark & \xmark & \cmark & \cmark & \cmark & \cmark & 15.520 ± 0.458 & 0.949 ± 0.023 & 1.000 ± 0.000 & 14.212 ± 1.496 & 8.659 ± 0.852 \\
\xmark & \xmark & \cmark & \cmark & \cmark & \cmark & \cmark & 97.324 ± 2.036 & 0.923 ± 0.037 & 1.000 ± 0.000 & 39.639 ± 0.999 & 16.273 ± 0.225 \\
\cmark & \cmark & \cmark & \xmark & \cmark & \cmark & \cmark & 17.496 ± 1.031 & 0.613 ± 0.121 & 0.715 ± 0.232 & 10.658 ± 1.695 & 7.278 ± 0.769 \\
\cmark & \cmark & \cmark & \cmark & \xmark & \cmark & \cmark & 18.883 ± 0.591 & 0.210 ± 0.070 & 0.048 ± 0.055 & 9.978 ± 1.017 & 6.360 ± 0.495 \\
\cmark & \cmark & \cmark & \cmark & \cmark & \xmark & \cmark & 19.070 ± 0.722 & 0.723 ± 0.035 & 1.000 ± 0.000 & 5.633 ± 1.501 & 3.449 ± 0.736 \\
\cmark & \cmark & \cmark & \cmark & \cmark & \xmark & \cmark & 18.812 ± 0.619 & 0.712 ± 0.052 & 1.000 ± 0.000 & 5.737 ± 1.484 & 3.579 ± 0.640 \\
\cmark & \cmark & \cmark & \xmark & \cmark & \cmark & \cmark & 18.722 ± 0.306 & 0.687 ± 0.042 & 0.985 ± 0.033 & 5.459 ± 1.241 & 3.574 ± 0.710 \\
\cmark & \cmark & \xmark & \cmark & \cmark & \cmark & \cmark & 17.774 ± 0.116 & 0.754 ± 0.048 & 1.000 ± 0.000 & 7.056 ± 1.288 & 4.182 ± 0.638 \\
\cmark & \cmark & \cmark & \cmark & \cmark & \cmark & \xmark & 20.643 ± 0.744 & 0.772 ± 0.047 & 1.000 ± 0.000 & 5.327 ± 0.850 & 2.791 ± 0.369 \\
\cmark & \cmark & \cmark & \cmark & \cmark & \xmark & \xmark & 20.563 ± 0.670 & 0.764 ± 0.053 & 1.000 ± 0.000 & 5.183 ± 0.432 & 2.792 ± 0.196 \\
\cmark & \xmark & \xmark & \cmark & \xmark & \cmark & \cmark & 23.091 ± 1.756 & 0.999 ± 0.001 & 1.000 ± 0.000 & 21.971 ± 2.942 & 11.986 ± 1.081 \\
\cmark & \xmark & \cmark & \cmark & \xmark & \cmark & \cmark & 15.520 ± 0.451 & 0.949 ± 0.024 & 1.000 ± 0.000 & 14.191 ± 1.490 & 8.659 ± 0.823 \\
\cmark & \cmark & \xmark & \cmark & \xmark & \cmark & \cmark & 97.324 ± 2.036 & 0.923 ± 0.037 & 1.000 ± 0.000 & 39.639 ± 0.999 & 16.273 ± 0.225 \\
\cmark & \xmark & \cmark & \cmark & \cmark & \cmark & \cmark & 15.540 ± 0.451 & 0.942 ± 0.023 & 1.000 ± 0.000 & 14.000 ± 1.615 & 8.523 ± 0.889 \\
\bottomrule

\end{tabular}
\end{subtable}
\end{table}

\begin{table}[H]
\small
\setlength{\tabcolsep}{5pt}
\renewcommand{\arraystretch}{1.15}
\captionsetup{labelformat=empty}
\caption{(c) GMSC}
\captionsetup{labelformat=default}
\centering
\begin{tabular}{cccccccccccc}
\toprule 
\textbf{\rot{Proximity}} & \textbf{\rot{Plaus\_p+}} & \textbf{\rot{Plaus\_p-}} & \textbf{\rot{Sparsity}} & \textbf{\rot{Validity}} & \textbf{\rot{Entropy}} & \textbf{\rot{PPT\_block}} & \textbf{NLL} & \textbf{$\hat{y}$} & \textbf{validity} & \textbf{similarity} & \textbf{sparsity} \\ 
\midrule 
\cmark & \cmark & \cmark & \cmark & \cmark & \cmark & \cmark & 18.838 ± 2.352 & 0.907 ± 0.085 & 0.975 ± 0.037 & 14.126 ± 1.751 & 5.242 ± 0.266 \\
\xmark & \cmark & \cmark & \cmark & \cmark & \cmark & \cmark & 23.481 ± 1.894 & 0.967 ± 0.038 & 1.000 ± 0.000 & 19.551 ± 2.491 & 7.319 ± 0.014 \\
\cmark & \xmark & \xmark & \cmark & \cmark & \cmark & \cmark & 17.323 ± 2.536 & 0.454 ± 0.270 & 0.706 ± 0.294 & 9.335 ± 2.814 & 4.495 ± 0.461 \\
\xmark & \cmark & \xmark & \cmark & \cmark & \cmark & \cmark & 11.304 ± 0.900 & 0.146 ± 0.032 & 0.957 ± 0.062 & 7.769 ± 0.365 & 4.741 ± 0.207 \\
\xmark & \xmark & \cmark & \cmark & \cmark & \cmark & \cmark & 47.875 ± 1.262 & 0.928 ± 0.089 & 1.000 ± 0.000 & 39.557 ± 1.529 & 7.903 ± 0.172 \\
\cmark & \cmark & \cmark & \xmark & \cmark & \cmark & \cmark & 17.779 ± 1.587 & 0.879 ± 0.128 & 1.000 ± 0.000 & 12.055 ± 1.159 & 5.135 ± 0.392 \\
\cmark & \cmark & \cmark & \cmark & \xmark & \cmark & \cmark & 14.493 ± 0.584 & 0.043 ± 0.019 & 0.184 ± 0.186 & 7.646 ± 0.742 & 4.463 ± 0.070 \\
\cmark & \cmark & \cmark & \cmark & \cmark & \xmark & \cmark & 14.417 ± 1.274 & 0.277 ± 0.132 & 1.000 ± 0.000 & 6.584 ± 1.100 & 2.729 ± 0.623 \\
\cmark & \cmark & \cmark & \cmark & \cmark & \xmark & \cmark & 14.738 ± 0.972 & 0.271 ± 0.108 & 1.000 ± 0.000 & 6.861 ± 1.211 & 2.597 ± 0.596 \\
\cmark & \cmark & \cmark & \xmark & \cmark & \cmark & \cmark & 14.819 ± 1.015 & 0.230 ± 0.098 & 1.000 ± 0.000 & 6.008 ± 0.845 & 2.347 ± 0.333 \\
\cmark & \cmark & \xmark & \cmark & \cmark & \cmark & \cmark & 11.943 ± 0.970 & 0.132 ± 0.014 & 0.997 ± 0.006 & 6.764 ± 0.390 & 3.513 ± 0.389 \\
\cmark & \cmark & \cmark & \cmark & \cmark & \cmark & \xmark & 20.555 ± 0.557 & 0.578 ± 0.115 & 0.944 ± 0.046 & 7.393 ± 0.697 & 1.868 ± 0.222 \\
\cmark & \cmark & \cmark & \cmark & \cmark & \xmark & \xmark & 18.194 ± 1.109 & 0.279 ± 0.087 & 0.781 ± 0.150 & 6.235 ± 0.582 & 2.294 ± 0.161 \\
\cmark & \xmark & \xmark & \cmark & \xmark & \cmark & \cmark & 16.163 ± 0.881 & 0.014 ± 0.004 & 0.038 ± 0.050 & 4.633 ± 0.501 & 3.812 ± 0.083 \\
\cmark & \xmark & \cmark & \cmark & \cmark & \cmark & \cmark & 47.821 ± 1.306 & 0.927 ± 0.088 & 1.000 ± 0.000 & 39.477 ± 1.526 & 7.855 ± 0.107 \\
\cmark & \cmark & \xmark & \cmark & \xmark & \cmark & \cmark & 11.339 ± 1.052 & 0.107 ± 0.046 & 0.760 ± 0.391 & 7.796 ± 0.849 & 4.674 ± 0.152 \\
\cmark & \xmark & \cmark & \cmark & \xmark & \cmark & \cmark & 47.915 ± 1.326 & 0.730 ± 0.329 & 0.998 ± 0.006 & 40.271 ± 0.487 & 7.762 ± 0.055 \\
\bottomrule 
\end{tabular}
\end{table}

\subsection{Robustness to Classifier Change}
\label{app:robustness_yhat}
\begin{figure*}[htbp]
  \centering
  \begin{subfigure}[t]{0.49\textwidth}
    \centering
    \includegraphics[width=\linewidth]{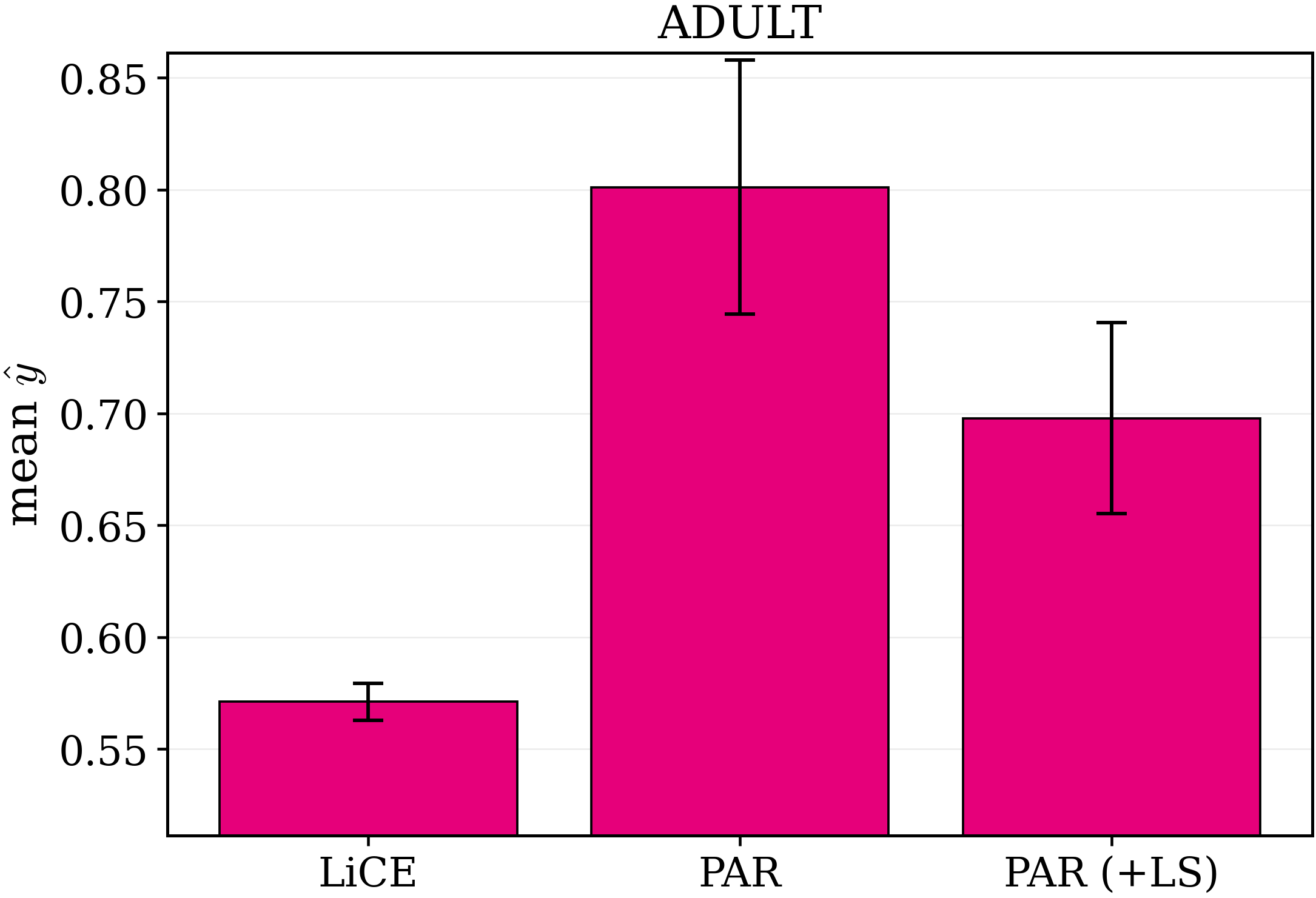}
    \caption{Adult}
    \label{fig:yhat-robust-adult}
  \end{subfigure}\hfill
  \begin{subfigure}[t]{0.49\textwidth}
    \centering
    \includegraphics[width=\linewidth]{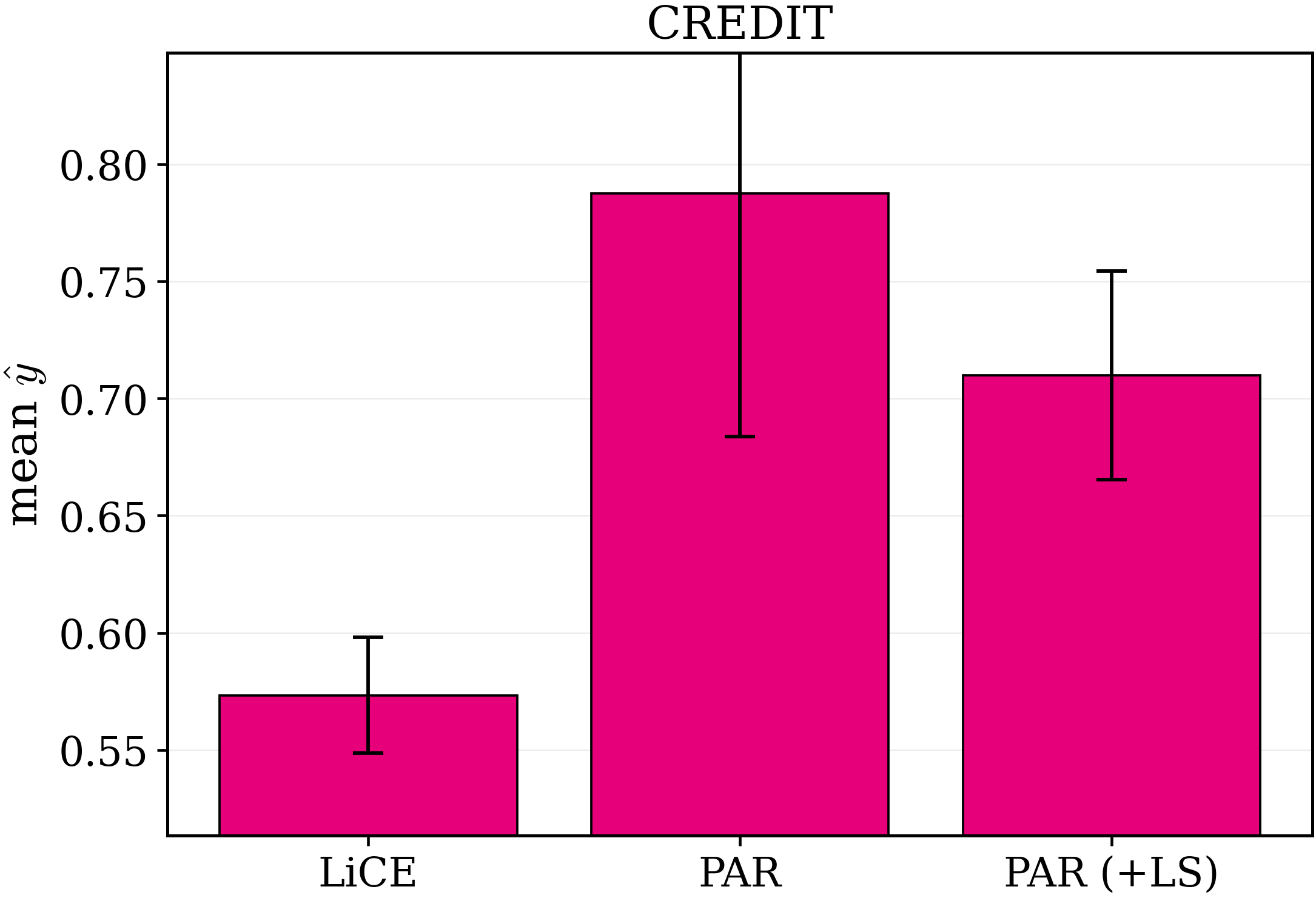}
    \caption{Credit}
    \label{fig:yhat-robust-credit}
  \end{subfigure}
  \caption{Robustness to classifier change measured by the mean predicted score $\hat{y}$ of generated counterfactuals under an alternative classifier (mean $\pm$ std across folds). Higher $\hat{y}$ indicates more stable decision flips under model replacement.}
  \label{fig:yhat-robustness}
\end{figure*}

We assess robustness to classifier change using the mean predicted score $\hat{y}$ of generated counterfactuals when evaluated under a classifier. 
Figure~\ref{fig:yhat-robustness} reports the mean $\hat{y}$ for LiCE and PAR, illustrating how sensitive each method’s counterfactuals are to changes in the decision boundary.

%% file: ijcai26.bib
@misc{nemecek2025generatinglikelycounterfactualsusing,
      title={Generating Likely Counterfactuals Using Sum-Product Networks}, 
      author={Jiri Nemecek and Tomas Pevny and Jakub Marecek},
      year={2025},
      eprint={2401.14086},
      archivePrefix={arXiv},
      primaryClass={cs.AI},
      url={https://arxiv.org/abs/2401.14086}, 
}

@inproceedings{St_pka_2025, series={KDD ’25},
   title={Counterfactual Explanations with Probabilistic Guarantees on their Robustness to Model Change},
   url={http://dx.doi.org/10.1145/3690624.3709300},
   DOI={10.1145/3690624.3709300},
   booktitle={Proceedings of the 31st ACM SIGKDD Conference on Knowledge Discovery and Data Mining V.1},
   publisher={ACM},
   author={Stępka, Ignacy and Stefanowski, Jerzy and Lango, Mateusz},
   year={2025},
   month=jul, pages={1277–1288},
   collection={KDD ’25} }

@misc{jiang2024provablyrobustplausiblecounterfactual,
      title={Provably Robust and Plausible Counterfactual Explanations for Neural Networks via Robust Optimisation}, 
      author={Junqi Jiang and Jianglin Lan and Francesco Leofante and Antonio Rago and Francesca Toni},
      year={2024},
      eprint={2309.12545},
      archivePrefix={arXiv},
      primaryClass={cs.LG},
      url={https://arxiv.org/abs/2309.12545}, 
}

@inproceedings{Pawelczyk_2020, series={WWW ’20},
   title={Learning Model-Agnostic Counterfactual Explanations for Tabular Data},
   url={http://dx.doi.org/10.1145/3366423.3380087},
   DOI={10.1145/3366423.3380087},
   booktitle={Proceedings of The Web Conference 2020},
   publisher={ACM},
   author={Pawelczyk, Martin and Broelemann, Klaus and Kasneci, Gjergji},
   year={2020},
   month=apr, pages={3126–3132},
   collection={WWW ’20} }

@inproceedings{Poyiadzi_2020, series={AIES ’20},
   title={FACE: Feasible and Actionable Counterfactual Explanations},
   url={http://dx.doi.org/10.1145/3375627.3375850},
   DOI={10.1145/3375627.3375850},
   booktitle={Proceedings of the AAAI/ACM Conference on AI, Ethics, and Society},
   publisher={ACM},
   author={Poyiadzi, Rafael and Sokol, Kacper and Santos-Rodriguez, Raul and De Bie, Tijl and Flach, Peter},
   year={2020},
   month=feb, pages={344–350},
   collection={AIES ’20} }

@article{Guidotti2022Counterfactual,
  author    = {Guidotti, Riccardo},
  title     = {Counterfactual explanations and how to find them: Literature review and benchmarking},
  journal   = {Data Mining and Knowledge Discovery},
  volume    = {36},
  number    = {5},
  pages     = {1993--2037},
  year      = {2022},
  month     = apr,
  publisher = {Springer},
  doi       = {10.1007/s10618-022-00831-6}
}

@article{Karimi2022AlgorithmicRecourse,
  author    = {Karimi, Amir-Hossein and Barthe, Gilles and Balle, Borja and Valera, Isabel},
  title     = {A Survey of Algorithmic Recourse: Contrastive Explanations and Consequential Recommendations},
  journal   = {ACM Computing Surveys},
  volume    = {55},
  number    = {5},
  pages     = {1--36},
  year      = {2022},
  publisher = {ACM},
  doi       = {10.1145/3494670}
}

@inproceedings{Ustun2019ActionableRecourse,
  author    = {Ustun, Berk and Spangher, Alexander and Liu, Yang},
  title     = {Actionable Recourse in Linear Classification},
  booktitle = {Proceedings of the Conference on Fairness, Accountability, and Transparency (FAT*)},
  year      = {2019},
  pages     = {10--19},
  publisher = {ACM},
  doi       = {10.1145/3287560.3287566}
}

@article{Wachter2018CounterfactualGDPR,
  author  = {Wachter, Sandra and Mittelstadt, Brent and Russell, Chris},
  title   = {Counterfactual Explanations Without Opening the Black Box: Automated Decisions and the GDPR},
  journal = {Harvard Journal of Law \& Technology},
  volume  = {31},
  number  = {2},
  pages   = {841--887},
  year    = {2018}
}

@inproceedings{Mothilal2020DiCE,
  author    = {Mothilal, Ramaravind Kommiya and Sharma, Amit and Tan, Chenhao},
  title     = {Explaining Machine Learning Classifiers through Diverse Counterfactual Explanations},
  booktitle = {Proceedings of the 2020 Conference on Fairness, Accountability, and Transparency (FAccT)},
  year      = {2020},
  pages     = {607--617},
  publisher = {ACM},
  doi       = {10.1145/3351095.3372850}
}

@article{VanLooveren2019ProtoCF,
  author  = {Van Looveren, Arnaud and Klaise, Janis},
  title   = {Interpretable Counterfactual Explanations Guided by Prototypes},
  journal = {arXiv preprint arXiv:1907.02584},
  year    = {2019}
}

@inproceedings{Joshi2019REVISE,
  author    = {Joshi, Shalmali and Koyejo, Oluwasanmi and Vijitbenjaronk, Warut and Kim, Been and Ghosh, Joydeep},
  title     = {REVISE: Towards Realistic Individual Recourse and Actionable Explanations in Black-Box Decision Making Systems},
  booktitle = {arXiv preprint},
  year      = {2019},
  note      = {arXiv:1907.09615}
}

@inproceedings{Kugelgen2022FairCausalRecourse,
  author    = {von K{\"u}gelgen, Julius and Karimi, Amir-Hossein and Bhatt, Umang and Valera, Isabel and Weller, Adrian and Sch{\"o}lkopf, Bernhard},
  title     = {On the Fairness of Causal Algorithmic Recourse},
  booktitle = {Proceedings of the AAAI Conference on Artificial Intelligence},
  volume    = {36},
  pages     = {9584--9594},
  year      = {2022},
  publisher = {AAAI Press}
}

@inproceedings{Choi2020ProbabilisticCircuits,
  author    = {Choi, YooJung and Vergari, Antonio and Van den Broeck, Guy},
  title     = {Probabilistic Circuits: A Unifying Framework for Tractable Probabilistic Models},
  booktitle = {Proceedings of the Thirty-Fifth Conference on Uncertainty in Artificial Intelligence (UAI)},
  year      = {2020}
}

@article{Darwiche2003Differential,
  author  = {Darwiche, Adnan},
  title   = {A Differential Approach to Inference in Bayesian Networks},
  journal = {Journal of the ACM},
  volume  = {50},
  number  = {3},
  pages   = {280--305},
  year    = {2003}
}

@article{Karimi2021CausalRecourse,
  author  = {Karimi, Amir-Hossein and Sch{\"o}lkopf, Bernhard and Valera, Isabel},
  title   = {Algorithmic Recourse under Imperfect Causal Knowledge: A Probabilistic Approach},
  journal = {NeurIPS},
  year    = {2021}
}

@article{Darwiche2020ProbabilisticCircuits,
  author  = {Darwiche, Adnan},
  title   = {Probabilistic Circuits: A Unifying Framework for Tractable Probabilistic Models},
  journal = {Journal of Artificial Intelligence Research},
  volume  = {67},
  pages   = {1--94},
  year    = {2020}
}

@article{Peharz2017SPN,
  author  = {Peharz, Robert and Gens, Robert and Pernkopf, Franz and Domingos, Pedro},
  title   = {On the Latent Variable Interpretation in Sum-Product Networks},
  journal = {IEEE Transactions on Pattern Analysis and Machine Intelligence},
  volume  = {39},
  number  = {10},
  pages   = {2030--2044},
  year    = {2017}
}

@misc{KaggleGiveMeSomeCredit,
  author       = {{Credit Fusion} and {Will Cukierski}},
  title        = {Give Me Some Credit},
  year         = {2011},
  howpublished = {\url{https://www.kaggle.com/competitions/GiveMeSomeCredit}},
  note         = {Kaggle competition}
}

@misc{BeckerKohavi1996Adult,
  author       = {Becker, Barry and Kohavi, Ronny},
  title        = {Adult},
  year         = {1996},
  howpublished = {UCI Machine Learning Repository},
  url          = {https://doi.org/10.24432/C5XW20},
  note         = {Accessed from UCI Machine Learning Repository}
}

@misc{Hofmann1994GermanCredit,
  author       = {Hofmann, Hans},
  title        = {Statlog (German Credit Data)},
  year         = {1994},
  howpublished = {UCI Machine Learning Repository},
  url          = {https://doi.org/10.24432/C5NC77},
  note         = {Accessed from UCI Machine Learning Repository}
}

@article{Wachter2017CounterfactualSSRN,
  author  = {Wachter, Sandra and Mittelstadt, Brent and Russell, Chris},
  title   = {Counterfactual Explanations Without Opening the Black Box: Automated Decisions and the GDPR},
  journal = {SSRN Electronic Journal},
  year    = {2017},
  issn    = {1556-5068},
  doi     = {10.2139/ssrn.3063289}
}

@misc{loconte2022deeprob,
  doi = {10.48550/ARXIV.2212.04403},
  url = {https://arxiv.org/abs/2212.04403},
  author = {Loconte, Lorenzo and Gala, Gennaro},
  title = {{DeeProb-kit}: a Python Library for Deep Probabilistic Modelling},
  publisher = {arXiv},
  year = {2022}
}

@inproceedings{Poon2011SPN,
  author    = {Poon, Hoifung and Domingos, Pedro},
  title     = {Sum-Product Networks: A New Deep Architecture},
  booktitle = {Proceedings of the 27th Conference on Uncertainty in Artificial Intelligence (UAI)},
  year      = {2011},
  pages     = {337--346}
}

@article{Youden1950Index,
  author  = {Youden, W. J.},
  title   = {Index for Rating Diagnostic Tests},
  journal = {Cancer},
  volume  = {3},
  number  = {1},
  pages   = {32--35},
  year    = {1950},
  doi     = {10.1002/1097-0142(1950)3:1<32::AID-CNCR2820030106>3.0.CO;2-3}
}
